\pdfoutput=1

\documentclass[11pt]{article}

\usepackage[preprint]{acl}

\usepackage{times}
\usepackage{latexsym}
\usepackage[ruled,norelsize]{algorithm2e}

\usepackage[T1]{fontenc}

\usepackage[utf8]{inputenc}

\usepackage{microtype}

\usepackage{inconsolata}

\usepackage{graphicx}
\usepackage{booktabs}
\usepackage{colortbl}
\usepackage{subcaption}
\usepackage{amsmath}
\usepackage{fdsymbol}
\usepackage{makecell}

\makeatletter
\DeclareRobustCommand\onedot{\futurelet\@let@token\@onedot}
\def\@onedot{\ifx\@let@token.\else.\null\fi\xspace}

\def\eg{\emph{e.g}\onedot}

\makeatother

\DeclareMathOperator{\LLM}{LLM}

\newcommand{\mycmdhelper}[3]{%
  \ifx*#1%
    \mathchoice{#3}{#3}{#3}{#3} 
  \else
    \ensuremath{#3}\xspace 
  \fi
}

\newcommand{\epmem}[1][]{\mycmdhelper{#1}{}{\mathcal{M}}}
\newcommand{\longmem}[1][]{\mycmdhelper{#1}{}{\mathcal{M}^{l}}}
\newcommand{\shortmem}[1][]{\mycmdhelper{#1}{}{\mathcal{M}^{s}}}
\newcommand{\gtshortmem}[1][]{\mycmdhelper{#1}{}{\mathcal{M}^{s}_{gt}}}
\newcommand{\genshortmem}[1][]{\mycmdhelper{#1}{}{\mathcal{M}^{s}_{gen}}}

\newcommand{\event}[3][]{%
  \if\relax\detokenize{#2}\relax
    \mycmdhelper{#1}{}{e_{#3}} 
  \else
    \mycmdhelper{#1}{}{e^{#2}_{\!#3}} 
  \fi
}
\newcommand{\narration}[3][]{%
  \if\relax\detokenize{#2}\relax
    \mycmdhelper{#1}{}{t_{#3}}
  \else
    \mycmdhelper{#1}{}{t^{#2}_{\!#3}}
  \fi
}
\newcommand{\gennarration}[3][]{%
  \if\relax\detokenize{#2}\relax
    \mycmdhelper{#1}{}{\hat t_{#3}}
  \else
    \mycmdhelper{#1}{}{\hat t^{#2}_{\!#3}}
  \fi
}
\newcommand{\video}[1][]{\mycmdhelper{#1}{}{\mathcal{E}}}
\newcommand{\data}[1][]{\mycmdhelper{#1}{}{\mathcal{D}}}

\newcommand{\opt}{OPT-2.7B\xspace}
\newcommand{\vicuna}{Vicuna-7B\xspace}

\newcommand{\cameo}{CAMEO\xspace}

\makeatother

\title{Memory Helps, but Confabulation Misleads:\\ Understanding Streaming Events in Videos with MLLMs}

\author{
 \textbf{Gengyuan Zhang\textsuperscript{$\spadesuit$,$\vardiamondsuit$}},
 \textbf{Mingcong Ding\textsuperscript{$\spadesuit$}},
 \textbf{Tong Liu\textsuperscript{$\spadesuit$,$\vardiamondsuit$}},
 \textbf{Yao Zhang\textsuperscript{$\spadesuit$,$\vardiamondsuit$}},
 \textbf{Volker Tresp\textsuperscript{$\spadesuit$,$\vardiamondsuit$}}
\\
 \textsuperscript{$\spadesuit$}Ludwig-Maximilians-Universität München, Germany \\
 \textsuperscript{$\vardiamondsuit$}Munich Center for Machine Learning, Germany
 \\
 \href{mailto:zhang@dbs.ifi.lmu.de}{{\tt zhang@dbs.ifi.lmu.de}}
}

\begin{document}
\maketitle
\begin{abstract}
Multimodal large language models (MLLMs) have demonstrated strong performance in understanding videos holistically, 
yet their ability to process streaming videos—videos are treated as a sequence of visual events—remains underexplored.
Intuitively, leveraging past events as memory can enrich contextual and temporal understanding of the current event.
In this paper, 
we show that leveraging memories as contexts helps MLLMs better understand video events.
However, because such memories rely on predictions of preceding events, they may contain misinformation, leading to confabulation and degraded performance.
To address this, we propose a confabulation-aware memory modification method that mitigates confabulated memory for memory-enhanced event understanding.
\end{abstract}

\section{Introduction}
Leveraging Multimodal Large Language Models (MLLMs) to understand events in videos \citep{yu2024eliciting,zhang2023video,maaz-etal-2024-video} has shown their effectiveness in a wide range of tasks due to the reasoning capabilities of LLMs.

Much of the existing research focuses on understanding holistic videos~\citep{zhang2023video,li2024llava}, where videos are treated as a single entity and MLLMs process them in one go. 
In real-world scenarios, however, videos are often streaming and multi-event, and events unfold sequentially.
Under such conditions, video events have significant temporal and contextual dependencies due to their inherent semantic relevance.
Akin to human cognition, knowledge of past memories can help in understanding the present effectively.

\begin{figure}
    \centering
    \includegraphics[width=\linewidth]{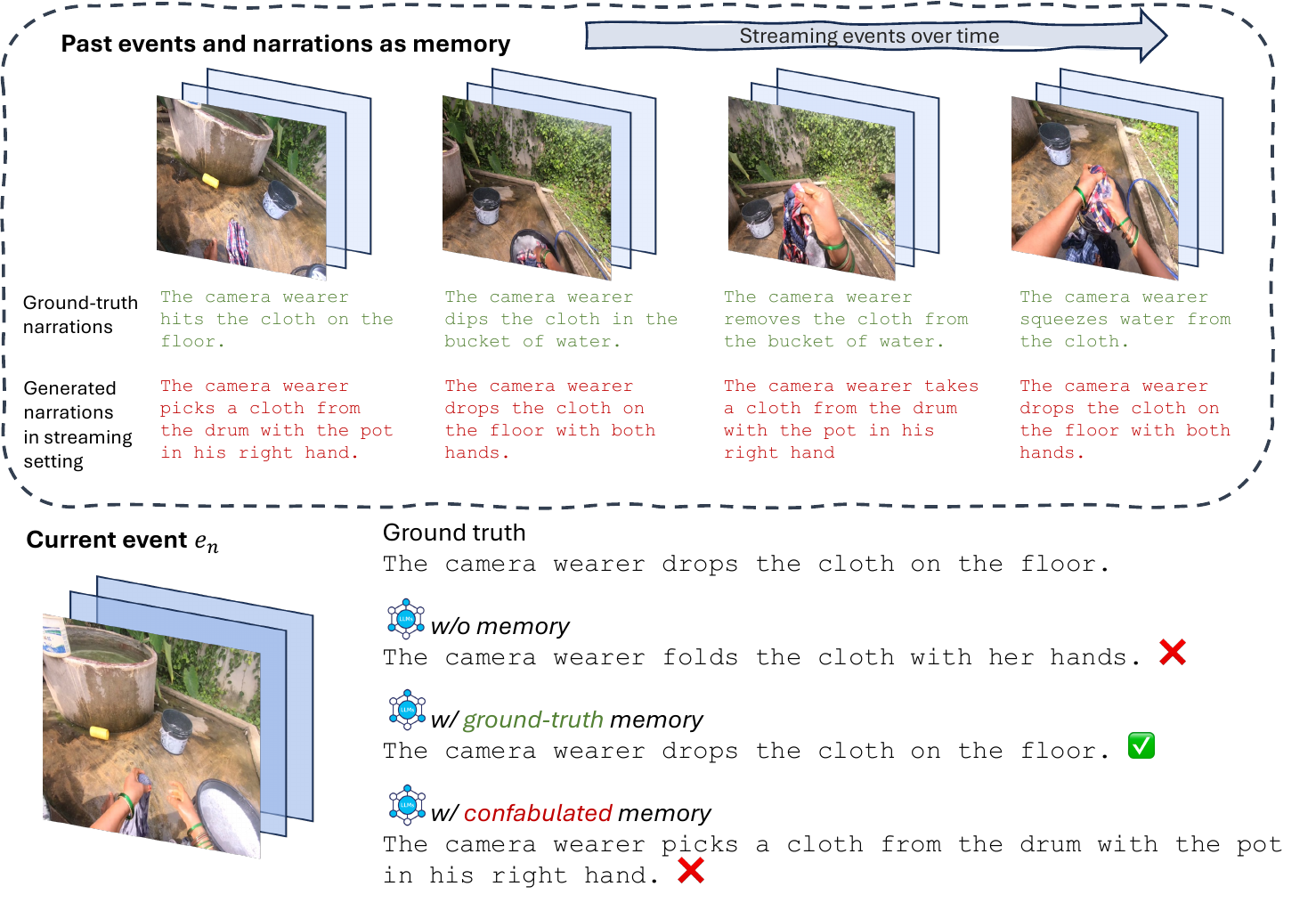}
    \caption{Knowing the memory of past events can help understand the current event. However, for streaming events in videos, we cannot access ground-truth narrations for previous events and this leads to confabulation.}
    \label{fig:teaser}
\end{figure}

We thus begin by exploring how incorporating memory mechanisms in LLMs can help enhance event understanding in videos.
A common practice for forming memory in LLMs is to prepend relevant contexts to the inputs of LLMs/MLLMs~\citep{behrouz2024titans,zhang2024survey,wang2024augmenting} where memories and knowledge are retained and recalled in tokens.  
Specifically, we specify the memory in the streaming video setting as event-triggered memory~\citep{fountas2024human,hatalis2023memory} since they are contextualized and dependent on the current scenes in videos, other than external event-agnostic knowledge.
The \textbf{memory} consists of \textit{long-term} memory~\citep{hatalis2023memory}, recollected from other episodes or videos with relevant events, and \textit{short-term} memory of events that precede the current one.

As shown in Fig.~\ref{fig:teaser} and later in Tab.\ref{tab:perf}, integrating memory as contexts, which consists of events and their narrations (detailed in Sec.~\ref{sec:method}), significantly outperforms the original 0-shot model, highlighting that knowing the past notably helps MLLMs.

Nevertheless, this performance represents a gold-standard baseline—feasible only in ideal scenarios rather than in a practical streaming setting.
In practice, MLLMs don't access the ground truth memory, rendering the leveraged short-term memory prone to mispredictions in the previous rounds due to hallucinations or incomplete observations. 
We reevaluate the model under streaming conditions and observe a substantial performance drop. We attribute this performance gap primarily to misinformation generated in the memory, a phenomenon referred to as \textbf{confabulation} \citep{sui-etal-2024-confabulation}.

To address the inevitable confabulated memory in the streaming setting, 
we introduce CAMEO, a confabulation-aware memory modification approach to mitigate the confabulation problem. 

Our paper can be summarized as follows:
\begin{enumerate}
    \setlength{\itemsep}{0pt}
    \setlength{\parskip}{0pt}
    \item We demonstrate that leveraging memory as context in MLLMs improves understanding of video events by introducing contextual and temporal knowledge.
    \item Yet ground-truth memory is not accessible in the real streaming setting since the memory suffers from misinformation in MLLMs' generation, namely confabulation;
    \item We propose CAMEO, a memory modification approach, to mitigate confabulated memory in MLLMs.
\end{enumerate}

\section{Leverage Memory as Context in Event Understanding}\label{sec:method}
\subsection{Preliminary}
Given a video episode \video consisting of a sequence of $N$ events $\{\event[*]{}{1},\dots,\event[*]{}{N}\}$ with an arbitrary event \event{}{i} in the episode as a short video clip consisting of multiple frames. 
LLMs need to predict the narration \gennarration{}{n} of the current query event \event{}{n} with only access to the preceding events \event{}{1:n-1}.

\paragraph{Memory} Memory~\citep{fountas2024human, das2024larimar} in LLMs, akin to human brains, refers to retrieving relevant events from past experiences. 
We denote the memory activated by the current event \event{}{n} as  \epmem. We adopt two types of memory: long-term memory \longmem from the set of other episodes as persistent memory \data; and short-term memory \shortmem, which respectively means the memories that occurred earlier and memories in the current episode. Memory as contexts \epmem is defined as a set of events recollected:
\vspace{-0.1cm}
\begin{equation}
\begin{aligned}
    \epmem &= \longmem \cup \shortmem,\\
    \longmem &= \{ (\event{l}{1}, \narration{l}{1}), \dots, (\event{l}{N_l}, \narration{l}{N_l}) \},\\
    \shortmem &= \{ (\event{s}{1}, \narration{s}{1}), \dots, (\event{s}{N_s}, \narration{s}{N_s}) \}.
\end{aligned}
\end{equation}
\vspace{-0.2cm}

To collect events \event[]{l}{j} in \longmem, we use a similarity-based retrieval method \citep{}. In contrast, for events \event[]{s}{k} in \shortmem, we rely on recency, selecting the most recent events relative to the current event (detailed in Sec.~\ref{sec:more-on-model}).

\paragraph{Memory as Context} A prominent trend in memory-enhanced LLMs is to use memory as contexts in LLMs’ inputs~\citep{fountas2024human, behrouz2024titans}, thereby leveraging the In-Context Learning (ICL) capability of LLMs~\citep{brown2020language,gao2023retrieval}. 
We formulate long-term and short-term memories as prepended contexts to the inputs.

\begin{figure*}[h]
    \centering
    \includegraphics[width=0.92\linewidth]{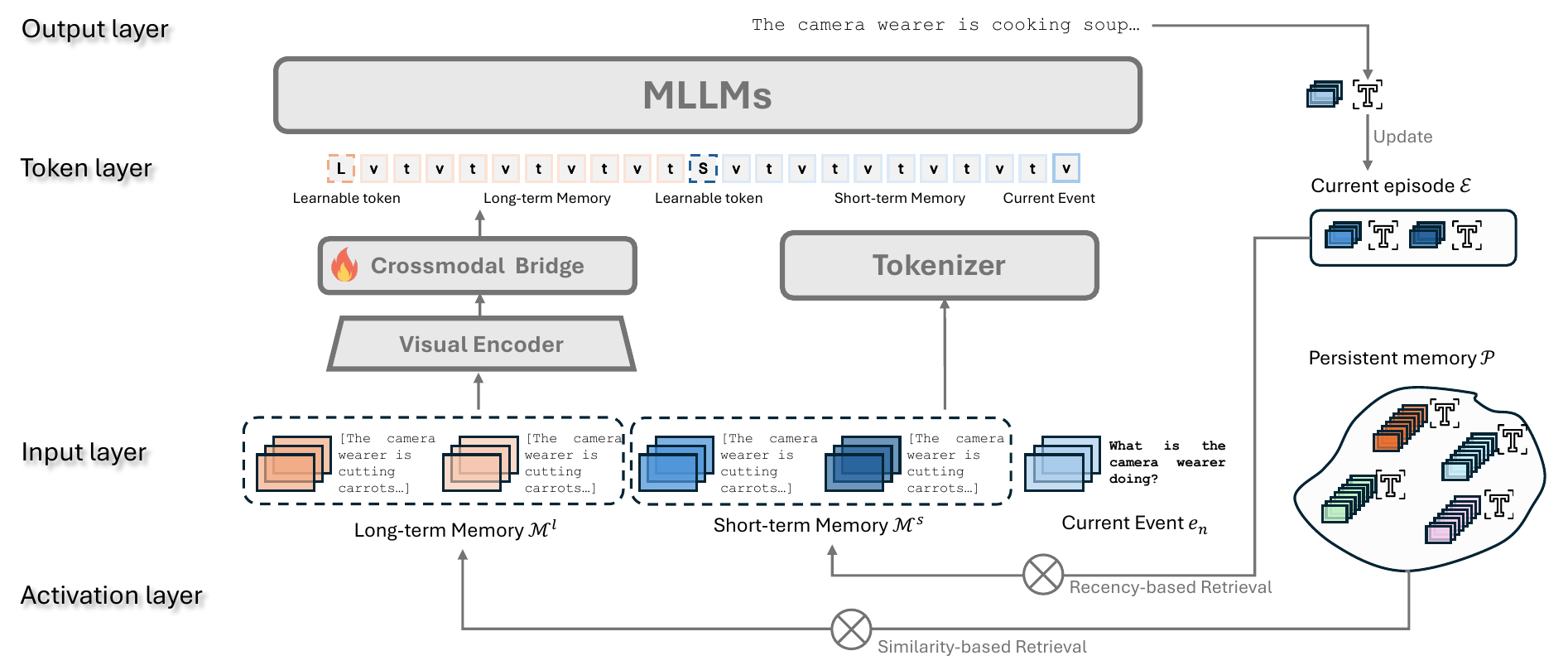}
    \caption{Model Pipeline for memory as contexts for streaming events reasoning. We interleave the events and narrations from long-term and short-term memory as contextual inputs.}
    \label{fig:enter-label}
\end{figure*}

\vspace{-0.2cm}
\begin{algorithm}
\caption{Streaming Evaluation}
\label{alg:training}
\KwIn{Current query event \event{}{n}, \\Current episode up to current event $\video_{:n-1}$, \\ Persistent memory \data}
\KwOut{Prediction of narration \narration{}{n}}
Initialize $\shortmem \gets \emptyset$\;
\For{$n = 1$ to $N$}{
Collect long-term memory from \data \\
\For{$k = 1$ to $N_l$}{
    $\longmem \gets (\event{l}{k}, \narration{}{k}) \in \data$\;
}
$\gennarration{}{n} \gets \LLM(\epmem, \event{}{n})$ \;
Update short-term memory with new prediction \\
$\shortmem \gets (\event{}{n}, \gennarration{}{n})$ \;
}
\Return \gennarration{}{n}\;
\end{algorithm}
\vspace{-0.6cm}

\subsection{Gain \& Loss: Memory in Event Understanding}
\paragraph{An Upper Bound: Ground-truth Memory helps}
We begin by introducing a golden baseline: we use \epmem with ground-truth narrations in \shortmem, denoted by \gtshortmem. 
In this setup, all memories are factual and plausible, therefore providing an upper bound on the performance of memory-enhanced LLMs.

\paragraph{Confabulated Memory Misleads Streaming Event Understanding}
We next evaluate LLMs’ event understanding capabilities in a streaming scenario, as illustrated in Alg.~\ref{alg:training}. 
This setting is more practical and challenging since the model cannot access ground-truth short-term memory \gtshortmem or rather the true narrations and can only utilize generated narrations up to the current event \genshortmem.

As shown in Fig.~\ref{fig:enter-label}, our streaming evaluation relies on short-term memory that incorporates earlier events and their predicted narrations.
Each new event’s prediction is subsequently consolidated into the short-term memory of the current episode.

\subsection{CAMEO: Mitigating Confabulation}
We propose a probing-and-surgical approach, \cameo, \textbf{C}onfabulation-\textbf{A}ware \textbf{ME}mory m\textbf{O}dification, to mitigate confabulation in our setting.
We treat confabulated memories as less factual and less credible contexts for LLMs.
Accordingly, we propose to quantize the credibility of short-term memory and reduce the contribution of heavily confabulated memories.

\paragraph{Uncertainty Estimation}
We adopt the semantic entropy proposed by \citet{kuhn2023semantic,farquhar2024detecting} as a proxy for assessing the credibility of generated narrations. Specifically, we sample $S$ generations of target narrations $\{\gennarration{(1)}{n},\dots,\gennarration{(S)}{n}\}$ from the output distribution of the LLMs conditioned on the memory \epmem and query event \event{}{n} from the posterior distribution $p(\gennarration{}{n} | \epmem[*], \event[*]{}{n}))$. 
\textit{Next}, we cluster the sampled narrations by their semantic equivalence, as described by \citet{kuhn2023semantic}, to identify the semantic clusters $C$ of generated narrations. The resulting semantic entropy is calculated as in Eq.\ref{eq:se}
\vspace{-0.2cm}
\begin{equation}
    se(\gennarration{}{n}) = -\sum_{c} p(c|\epmem, \event{}{n}) \log p(c | \epmem[*], \event[*]{}{n}))
\label{eq:se}
\end{equation}
\vspace{-0.6cm}

\paragraph{Probing Confabulation-Prone Heads} 
Attention heads have been shown to represent different subspaces of contextual information in LLMs \citep{deng2024cram, elhage2021mathematical, meng2022locating} and \citep{deng2024cram} demonstrates that modifying attention heads can mitigate misinformation in textual contexts.

Inspired by \citep{deng2024cram}, we aim to locate the confabulation-prone attention heads in the transformer architecture. 
To do so, we manipulate one ground-truth narration \(\narration{}{j}\) with a random, irrelevant narration \(\narration{-}{j}\) in \(\data\), ablate an arbitrary attention head h in the transformer, and then calculate the output logits to quantify the indirect effect (IE) \citep{meng2022locating, deng2024cram}, thereby identifying the influence of this attention head.

\paragraph{Confabulation-aware Memory Modification} 
Once we identify the confabulation-prone heads, we can reweight the attention in the most influential attention heads to the confabulated memory \gennarration{}{n} them according to their semantic entropy, as detailed in Appx.~\ref{sec:cameo-more}. 
We convert semantic entropy into weights as follows:
\vspace{-0.2cm}
\begin{equation}
    w(\gennarration{}{n}) = 1/exp(-\tau \times se(\gennarration{}{n}))
\vspace{-0.2cm}
\end{equation}
where $\tau$ is a temperature coefficient, with which we can tweak the modification scale.

\section{Experimental Study}
We elaborate the model design (Sec.~\ref{sec:model}), implementation details(Sec.~\ref{sec:imple}), and results (Sec.~\ref{sec:result}).

\subsection{Model Design}\label{sec:model}
\paragraph{Model Architecture}
Our models follow common MLLM paradigm \citep{liu2023llava,instructblip,liu2023improvedllava,li2023blip}: a visual encoder, a cross-modal bridge, and an LLM backbone.
Events, treated as short video clips, can vary in length, so we downsample them to a fixed length $l_e=8$. We then encode each frame individually with the CLIP~\citep{radford2021learning} visual encoder.
Subsequently, we use an event-aware bridge to encode the events and align them to the language space of LLMs, as shown in Fig.~\ref{fig:enter-label}. More details can be found in Appx.~\ref{sec:more-on-model}.

By encoding all the events in the memory \epmem, we interleave these events with their narrations and provide them as context to the model, alongside the new query event and the question.

\paragraph{Model Selection} We employ two widely used Large Language Models in our experiments: \opt \citep{zhang2022opt} and \vicuna \citep{zheng2023judging}.
We also use the Q-Former proposed by \citet{li2023blip} as an effective bridge to compress visual features from clips. 
Although MLP-based projection \citep{liu2023improvedllava, liu2023llava} is a popular method for adapting vision to the language space, it increases computational overhead when processing multiple video inputs, making compression essential.
We initialize the Q-Former for \opt from \citet{instructblip} and for \vicuna from \citet{zhuminigpt}, then unfreeze each with training data.


\begin{table}
    \scriptsize 
    \centering
    \begin{tabular}{lccc}
    \toprule
        Model &  STS &  Rouge-L & BLEU\\
    \midrule
        \rowcolor{gray!20} \opt \\
        w/o memory & 0.495 & 0.520 & 0.179 \\
        $\Delta$    & \textcolor{green}{+0.160} & \textcolor{green}{+0.103} & \textcolor{green}{+0.167} \\
        w/ ground-truth memory (16 shots) & 0.655 & 0.623 & 0.346 \\
        $\Delta$   & \textcolor{red}{-0.152} & \textcolor{red}{-0.082} & \textcolor{red}{-0.163} \\
        w/ confabulated memory (16 shots)  & 0.503 & 0.541 & 0.183 \\
    \midrule
        \rowcolor{gray!20} \vicuna \\
        w/o memory & 0.615 & 0.588 & 0.240 \\
        $\Delta$   & \textcolor{green}{+0.178} & \textcolor{green}{+0.145} & \textcolor{green}{+0.293} \\
        w/ ground-truth memory (16 shots) & 0.793 & 0.733 & 0.534 \\
        $\Delta$   & \textcolor{red}{-0.137} & \textcolor{red}{-0.142} & \textcolor{red}{-0.220} \\
        w/ confabulated memory (16 shots) & 0.656 & 0.591 & 0.315 \\
    \bottomrule
    \end{tabular}
    \caption{Model Performance. Memory enhances event understanding by providing contextual knowledge of the current event, while confabulated memory in the streaming setting can mislead MLLMs.}
    \label{tab:perf}
\vspace{-0.4cm}
\end{table}

\subsection{Implementation}\label{sec:imple}
\paragraph{Training Details}
To elicit the multimodal in-context learning capability of MLLMs, we retrain them with memory \epmem as interleaved video-text data \citep{yu2024eliciting, alayrac2022flamingo, li2024textbind, wang2024cosmo}, following Alg.\ref{alg:training} in Appx.~\ref{sec:training}.
We pad 8 events sampled from \(\data\) and 8 most recent events in the current episode \video and adopt two new learnable tokens to distinguish different memories. The total shot number is 16.
We unfreeze the Q-Former to adapt to visual events.

\paragraph{Dataset}
We use the Ego4D dataset \citep{grauman2022ego4d} for our task. To construct \data from past episodes and for training, we utilize the training split. To probe the confabulation-prone heads, we rely on the validation set. Finally, for all evaluations, we employ the test set.

\paragraph{Evaluation}
We evaluate the narrations with Semantic Textual Similarity (STS) by Sentence-BERT~\citet{reimers-gurevych-2019-sentence}, ROUGE-L by \citet{lin-2004-rouge}, BLEU by \citet{papineni2002BLEU}.

\subsection{Results}\label{sec:result}

\paragraph{Memory helps, but confabulation misleads}
We first compare 3 baselines: MLLMs without memory, MLLMs with ground-truth memory (w/ ground-truth narrations), and MLLMs with confabulated memory (w/ generated narrations) in Tab.~\ref{tab:perf}.

First, we observe a significant performance gain when MLLMs leverage memory as contexts, yielding a large margin of improvement across all mentioned metrics for both models.

However, the performance drop in streaming evaluation is concerning. Its primary cause is that in the streaming scenario, we replace ground-truth narrations with previously generated narrations of recent events. This inevitably leads to a performance drop, stemming from the confabulation introduced by earlier predictions.

\begin{figure}[ht]
    \centering
    
    \begin{subfigure}[b]{0.49\linewidth}
        \centering
        \includegraphics[width=\textwidth]{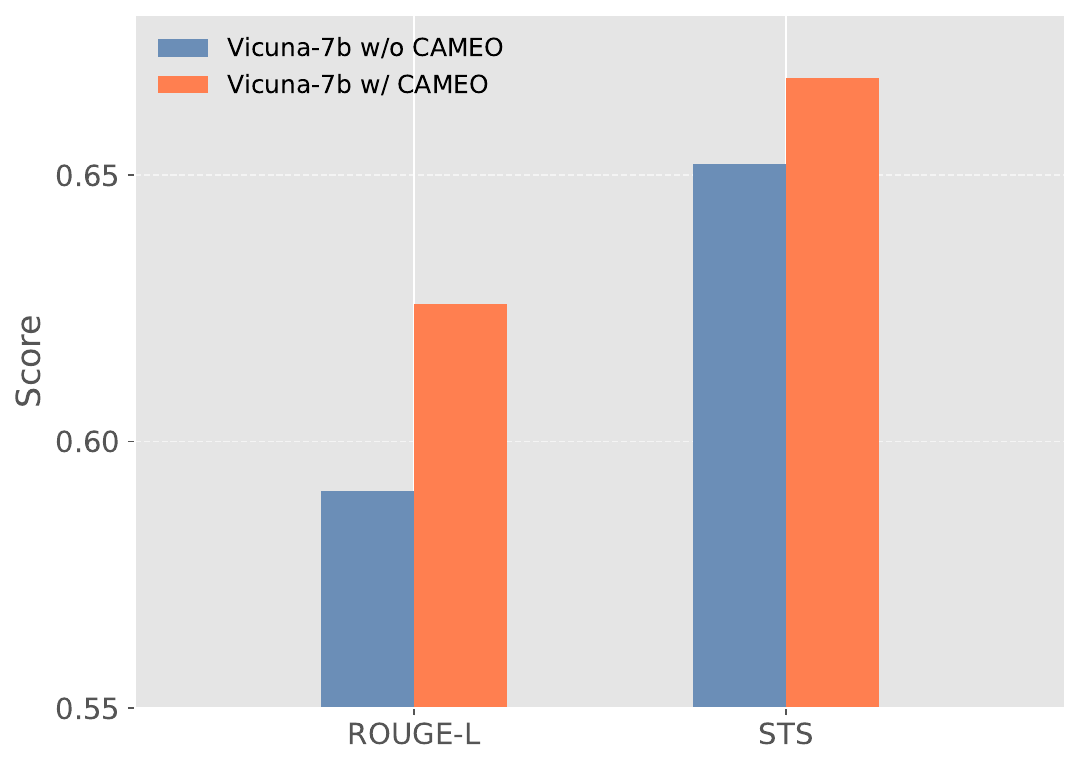} 
        \caption{\vicuna}
        \label{fig:mod-vicuna}
    \end{subfigure}
    \hfill
    \begin{subfigure}[b]{0.49\linewidth}
        \centering
        \includegraphics[width=\textwidth]{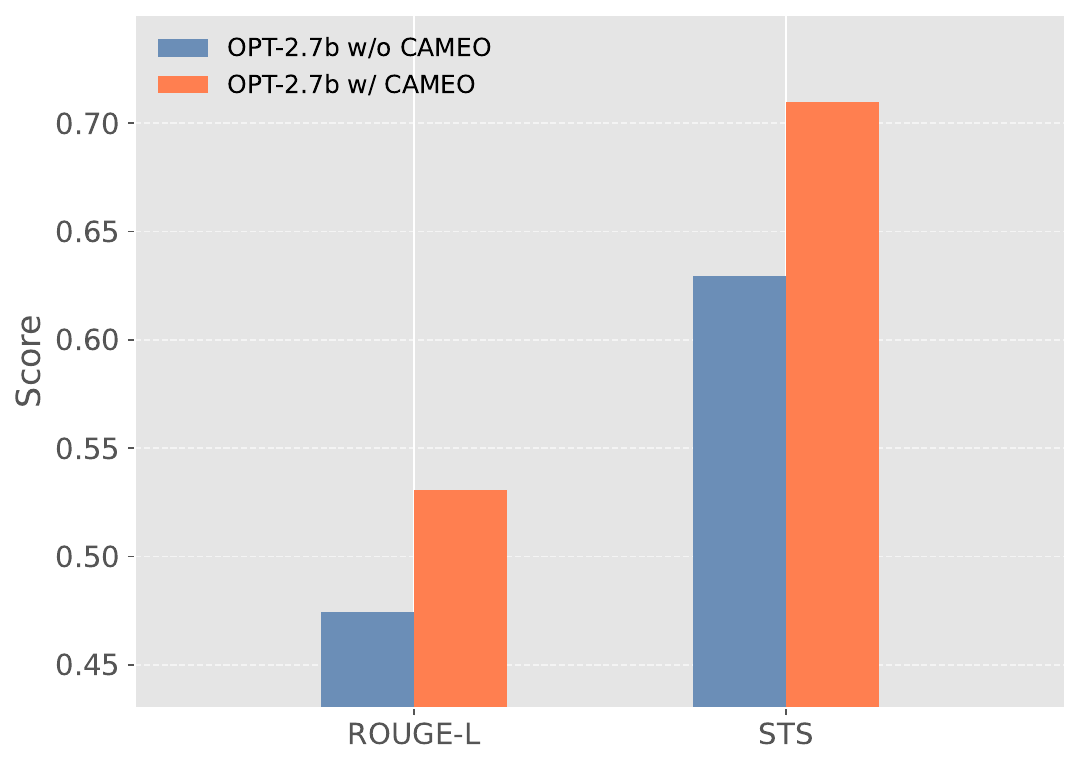} 
        \caption{\opt}
        \label{fig:mod-opt}
    \end{subfigure}

    \caption{Performance improvements with \cameo.}
    \label{fig:mod}
\end{figure}

\paragraph{Mitigating Confabulation}
As shown in Fig.~\ref{fig:mod}, with CAMEO, the reasoning capabilities of MLLMs for streaming event understanding improve significantly. This result underscores the effectiveness of CAMEO in combating confabulated memory.

\section{Conclusion}
In this work, we demonstrated how memory, leveraged as context, can inject temporal knowledge into MLLMs and enhance their understanding of visual events in streaming videos. However, in practical streaming scenarios, accessing ground-truth memory is often impossible, leading to confabulation when predictions depend on previously generated narrations. We addressed this issue by proposing a confabulation-aware attention modification mechanism, which uses the semantic uncertainty of predicted narrations as a proxy for the credibility of potentially confabulation-prone memory. We show that this approach effectively mitigates confabulated memory.

\paragraph{Limitations}
\begin{enumerate}
    \item In this work, we focus on forming memory as context for MLLMs. While this approach is training-free and allows us to exploit MLLMs’ capabilities fully, future efforts could investigate parameterized memory components, as discussed by \citet{behrouz2024titans}.
    \item Our current setting relies on a predefined semantic boundary between events in the original dataset to simplify the task. Thus, videos are streamed at the event level rather than the frame level. A more realistic streaming scenario would be more challenging because the semantic boundary is unknown.
    \item Due to computational constraints, our model selection is limited. Implementing interleaved video-text inputs requires a powerful architecture to compress visual features and reduce token length (\eg, Q-Former). This hinders our ability to explore other models that do not compress visual features, as their computational complexity grows quadratically with the amount of memory shots.
\end{enumerate}

\paragraph{Ethics Statement}
In this work, we investigate the cause and effect of MLLMs' confabulation in streaming event understanding for videos. Such confabulation is caused by MLLMs' mispredictions, may reflect intrinsic hallucination, and lead to misinformation, resulting in unsolicited or undesired outputs.


\bibliography{custom}

\begin{thebibliography}{39}
\providecommand{\natexlab}[1]{#1}

\bibitem[{Alayrac et~al.(2022)Alayrac, Donahue, Luc, Miech, Barr, Hasson, Lenc, Mensch, Millican, Reynolds et~al.}]{alayrac2022flamingo}
Jean-Baptiste Alayrac, Jeff Donahue, Pauline Luc, Antoine Miech, Iain Barr, Yana Hasson, Karel Lenc, Arthur Mensch, Katherine Millican, Malcolm Reynolds, et~al. 2022.
\newblock Flamingo: a visual language model for few-shot learning.
\newblock \emph{Advances in neural information processing systems}, 35:23716--23736.

\bibitem[{Behrouz et~al.(2024)Behrouz, Zhong, and Mirrokni}]{behrouz2024titans}
Ali Behrouz, Peilin Zhong, and Vahab Mirrokni. 2024.
\newblock Titans: Learning to memorize at test time.
\newblock \emph{arXiv preprint arXiv:2501.00663}.

\bibitem[{Brown et~al.(2020)Brown, Mann, Ryder, Subbiah, Kaplan, Dhariwal, Neelakantan, Shyam, Sastry, Askell et~al.}]{brown2020language}
Tom Brown, Benjamin Mann, Nick Ryder, Melanie Subbiah, Jared~D Kaplan, Prafulla Dhariwal, Arvind Neelakantan, Pranav Shyam, Girish Sastry, Amanda Askell, et~al. 2020.
\newblock Language models are few-shot learners.
\newblock \emph{Advances in neural information processing systems}, 33:1877--1901.

\bibitem[{Chen et~al.(2024)Chen, Zhang, Han, He, and Li}]{chen2024camml}
Yixin Chen, Shuai Zhang, Boran Han, Tong He, and Bo~Li. 2024.
\newblock \href {https://doi.org/10.18653/v1/2024.acl-long.223} {{C}a{MML}: Context-aware multimodal learner for large models}.
\newblock In \emph{Proceedings of the 62nd Annual Meeting of the Association for Computational Linguistics (Volume 1: Long Papers)}, pages 4056--4071, Bangkok, Thailand. Association for Computational Linguistics.

\bibitem[{Dai et~al.(2023)Dai, Li, Li, Tiong, Zhao, Wang, Li, Fung, and Hoi}]{instructblip}
Wenliang Dai, Junnan Li, Dongxu Li, Anthony Meng~Huat Tiong, Junqi Zhao, Weisheng Wang, Boyang Li, Pascale Fung, and Steven Hoi. 2023.
\newblock \href {https://arxiv.org/abs/2305.06500} {Instructblip: Towards general-purpose vision-language models with instruction tuning}.
\newblock \emph{Preprint}, arXiv:2305.06500.

\bibitem[{Das et~al.(2024)Das, Chaudhury, Nelson, Melnyk, Swaminathan, Dai, Lozano, Kollias, Chenthamarakshan, Dan et~al.}]{das2024larimar}
Payel Das, Subhajit Chaudhury, Elliot Nelson, Igor Melnyk, Sarath Swaminathan, Sihui Dai, Aur{\'e}lie Lozano, Georgios Kollias, Vijil Chenthamarakshan, Soham Dan, et~al. 2024.
\newblock Larimar: Large language models with episodic memory control.
\newblock \emph{arXiv preprint arXiv:2403.11901}.

\bibitem[{Deng et~al.(2024)Deng, Wang, Zhu, Wang, and Feng}]{deng2024cram}
Boyi Deng, Wenjie Wang, Fengbin Zhu, Qifan Wang, and Fuli Feng. 2024.
\newblock Cram: Credibility-aware attention modification in llms for combating misinformation in rag.
\newblock \emph{arXiv preprint arXiv:2406.11497}.

\bibitem[{Elhage et~al.(2021)Elhage, Nanda, Olsson, Henighan, Joseph, Mann, Askell, Bai, Chen, Conerly et~al.}]{elhage2021mathematical}
Nelson Elhage, Neel Nanda, Catherine Olsson, Tom Henighan, Nicholas Joseph, Ben Mann, Amanda Askell, Yuntao Bai, Anna Chen, Tom Conerly, et~al. 2021.
\newblock A mathematical framework for transformer circuits.
\newblock \emph{Transformer Circuits Thread}, 1(1):12.

\bibitem[{Fan et~al.(2024)Fan, Ma, Wu, Du, Li, Gao, and Li}]{fan2024videoagent}
Yue Fan, Xiaojian Ma, Rujie Wu, Yuntao Du, Jiaqi Li, Zhi Gao, and Qing Li. 2024.
\newblock Videoagent: A memory-augmented multimodal agent for video understanding.
\newblock In \emph{European Conference on Computer Vision}, pages 75--92. Springer.

\bibitem[{Farquhar et~al.(2024)Farquhar, Kossen, Kuhn, and Gal}]{farquhar2024detecting}
Sebastian Farquhar, Jannik Kossen, Lorenz Kuhn, and Yarin Gal. 2024.
\newblock Detecting hallucinations in large language models using semantic entropy.
\newblock \emph{Nature}, 630(8017):625--630.

\bibitem[{Fountas et~al.(2025)Fountas, Benfeghoul, Oomerjee, Christopoulou, Lampouras, Ammar, and Wang}]{fountas2024human}
Zafeirios Fountas, Martin Benfeghoul, Adnan Oomerjee, Fenia Christopoulou, Gerasimos Lampouras, Haitham~Bou Ammar, and Jun Wang. 2025.
\newblock \href {https://openreview.net/forum?id=BI2int5SAC} {Human-like episodic memory for infinite context {LLM}s}.
\newblock In \emph{The Thirteenth International Conference on Learning Representations}.

\bibitem[{Gao et~al.(2023)Gao, Xiong, Gao, Jia, Pan, Bi, Dai, Sun, and Wang}]{gao2023retrieval}
Yunfan Gao, Yun Xiong, Xinyu Gao, Kangxiang Jia, Jinliu Pan, Yuxi Bi, Yi~Dai, Jiawei Sun, and Haofen Wang. 2023.
\newblock Retrieval-augmented generation for large language models: A survey.
\newblock \emph{arXiv preprint arXiv:2312.10997}.

\bibitem[{Grauman et~al.(2022)Grauman, Westbury, Byrne, Chavis, Furnari, Girdhar, Hamburger, Jiang, Liu, Liu et~al.}]{grauman2022ego4d}
Kristen Grauman, Andrew Westbury, Eugene Byrne, Zachary Chavis, Antonino Furnari, Rohit Girdhar, Jackson Hamburger, Hao Jiang, Miao Liu, Xingyu Liu, et~al. 2022.
\newblock Ego4d: Around the world in 3,000 hours of egocentric video.
\newblock In \emph{Proceedings of the IEEE/CVF Conference on Computer Vision and Pattern Recognition}, pages 18995--19012.

\bibitem[{Hatalis et~al.(2023)Hatalis, Christou, Myers, Jones, Lambert, Amos-Binks, Dannenhauer, and Dannenhauer}]{hatalis2023memory}
Kostas Hatalis, Despina Christou, Joshua Myers, Steven Jones, Keith Lambert, Adam Amos-Binks, Zohreh Dannenhauer, and Dustin Dannenhauer. 2023.
\newblock Memory matters: The need to improve long-term memory in llm-agents.
\newblock In \emph{Proceedings of the AAAI Symposium Series}, volume~2, pages 277--280.

\bibitem[{He et~al.(2024)He, Li, Jang, Jia, Cao, Shah, Shrivastava, and Lim}]{he2024ma}
Bo~He, Hengduo Li, Young~Kyun Jang, Menglin Jia, Xuefei Cao, Ashish Shah, Abhinav Shrivastava, and Ser-Nam Lim. 2024.
\newblock Ma-lmm: Memory-augmented large multimodal model for long-term video understanding.
\newblock In \emph{Proceedings of the IEEE/CVF Conference on Computer Vision and Pattern Recognition}, pages 13504--13514.

\bibitem[{Kuhn et~al.(2023)Kuhn, Gal, and Farquhar}]{kuhn2023semantic}
Lorenz Kuhn, Yarin Gal, and Sebastian Farquhar. 2023.
\newblock \href {https://openreview.net/forum?id=VD-AYtP0dve} {Semantic uncertainty: Linguistic invariances for uncertainty estimation in natural language generation}.
\newblock In \emph{The Eleventh International Conference on Learning Representations}.

\bibitem[{Li et~al.(2024{\natexlab{a}})Li, Zhang, Guo, Zhang, Li, Zhang, Zhang, Li, Liu, and Li}]{li2024llava}
Bo~Li, Yuanhan Zhang, Dong Guo, Renrui Zhang, Feng Li, Hao Zhang, Kaichen Zhang, Yanwei Li, Ziwei Liu, and Chunyuan Li. 2024{\natexlab{a}}.
\newblock Llava-onevision: Easy visual task transfer.
\newblock \emph{arXiv preprint arXiv:2408.03326}.

\bibitem[{Li et~al.(2024{\natexlab{b}})Li, Li, Cai, Wang, Liu, Watanabe, Yang, and Shi}]{li2024textbind}
Huayang Li, Siheng Li, Deng Cai, Longyue Wang, Lemao Liu, Taro Watanabe, Yujiu Yang, and Shuming Shi. 2024{\natexlab{b}}.
\newblock Textbind: Multi-turn interleaved multimodal instruction-following in the wild.
\newblock In \emph{Findings of the Association for Computational Linguistics ACL 2024}, pages 9053--9076.

\bibitem[{Li et~al.(2023)Li, Li, Savarese, and Hoi}]{li2023blip}
Junnan Li, Dongxu Li, Silvio Savarese, and Steven Hoi. 2023.
\newblock Blip-2: Bootstrapping language-image pre-training with frozen image encoders and large language models.
\newblock In \emph{International conference on machine learning}, pages 19730--19742. PMLR.

\bibitem[{Lin(2004)}]{lin-2004-rouge}
Chin-Yew Lin. 2004.
\newblock \href {https://aclanthology.org/W04-1013/} {{ROUGE}: A package for automatic evaluation of summaries}.
\newblock In \emph{Text Summarization Branches Out}, pages 74--81, Barcelona, Spain. Association for Computational Linguistics.

\bibitem[{Liu et~al.(2024)Liu, Li, Li, and Lee}]{liu2023improvedllava}
Haotian Liu, Chunyuan Li, Yuheng Li, and Yong~Jae Lee. 2024.
\newblock Improved baselines with visual instruction tuning.
\newblock In \emph{Proceedings of the IEEE/CVF Conference on Computer Vision and Pattern Recognition}, pages 26296--26306.

\bibitem[{Liu et~al.(2023)Liu, Li, Wu, and Lee}]{liu2023llava}
Haotian Liu, Chunyuan Li, Qingyang Wu, and Yong~Jae Lee. 2023.
\newblock Visual instruction tuning.
\newblock \emph{Advances in neural information processing systems}, 36:34892--34916.

\bibitem[{Maaz et~al.(2024)Maaz, Rasheed, Khan, and Khan}]{maaz-etal-2024-video}
Muhammad Maaz, Hanoona Rasheed, Salman Khan, and Fahad Khan. 2024.
\newblock \href {https://doi.org/10.18653/v1/2024.acl-long.679} {Video-{C}hat{GPT}: Towards detailed video understanding via large vision and language models}.
\newblock In \emph{Proceedings of the 62nd Annual Meeting of the Association for Computational Linguistics (Volume 1: Long Papers)}, pages 12585--12602, Bangkok, Thailand. Association for Computational Linguistics.

\bibitem[{Meng et~al.(2022)Meng, Bau, Andonian, and Belinkov}]{meng2022locating}
Kevin Meng, David Bau, Alex Andonian, and Yonatan Belinkov. 2022.
\newblock Locating and editing factual associations in gpt.
\newblock \emph{Advances in Neural Information Processing Systems}, 35:17359--17372.

\bibitem[{Papineni et~al.(2002)Papineni, Roukos, Ward, and Zhu}]{papineni2002BLEU}
Kishore Papineni, Salim Roukos, Todd Ward, and Wei-Jing Zhu. 2002.
\newblock Bleu: a method for automatic evaluation of machine translation.
\newblock In \emph{Proceedings of the 40th annual meeting of the Association for Computational Linguistics}, pages 311--318.

\bibitem[{Qian et~al.(2025)Qian, Dong, Zhang, Zang, Ding, Lin, and Wang}]{qian2025streaming}
Rui Qian, Xiaoyi Dong, Pan Zhang, Yuhang Zang, Shuangrui Ding, Dahua Lin, and Jiaqi Wang. 2025.
\newblock Streaming long video understanding with large language models.
\newblock \emph{Advances in Neural Information Processing Systems}, 37:119336--119360.

\bibitem[{Radford et~al.(2021)Radford, Kim, Hallacy, Ramesh, Goh, Agarwal, Sastry, Askell, Mishkin, Clark et~al.}]{radford2021learning}
Alec Radford, Jong~Wook Kim, Chris Hallacy, Aditya Ramesh, Gabriel Goh, Sandhini Agarwal, Girish Sastry, Amanda Askell, Pamela Mishkin, Jack Clark, et~al. 2021.
\newblock Learning transferable visual models from natural language supervision.
\newblock In \emph{International conference on machine learning}, pages 8748--8763. PMLR.

\bibitem[{Reimers and Gurevych(2019)}]{reimers-gurevych-2019-sentence}
Nils Reimers and Iryna Gurevych. 2019.
\newblock \href {https://doi.org/10.18653/v1/D19-1410} {Sentence-{BERT}: Sentence embeddings using {S}iamese {BERT}-networks}.
\newblock In \emph{Proceedings of the 2019 Conference on Empirical Methods in Natural Language Processing and the 9th International Joint Conference on Natural Language Processing (EMNLP-IJCNLP)}, pages 3982--3992, Hong Kong, China. Association for Computational Linguistics.

\bibitem[{Sui et~al.(2024)Sui, Duede, Wu, and So}]{sui-etal-2024-confabulation}
Peiqi Sui, Eamon Duede, Sophie Wu, and Richard So. 2024.
\newblock \href {https://doi.org/10.18653/v1/2024.acl-long.770} {Confabulation: The surprising value of large language model hallucinations}.
\newblock In \emph{Proceedings of the 62nd Annual Meeting of the Association for Computational Linguistics (Volume 1: Long Papers)}, pages 14274--14284, Bangkok, Thailand. Association for Computational Linguistics.

\bibitem[{Wang et~al.(2024{\natexlab{a}})Wang, Li, Lin, Wang, Lin, Yang, Wang, and Shou}]{wang2024cosmo}
Alex~Jinpeng Wang, Linjie Li, Kevin~Qinghong Lin, Jianfeng Wang, Kevin Lin, Zhengyuan Yang, Lijuan Wang, and Mike~Zheng Shou. 2024{\natexlab{a}}.
\newblock Cosmo: Contrastive streamlined multimodal model with interleaved pre-training.
\newblock \emph{arXiv preprint arXiv:2401.00849}.

\bibitem[{Wang et~al.(2024{\natexlab{b}})Wang, Dong, Cheng, Liu, Yan, Gao, and Wei}]{wang2024augmenting}
Weizhi Wang, Li~Dong, Hao Cheng, Xiaodong Liu, Xifeng Yan, Jianfeng Gao, and Furu Wei. 2024{\natexlab{b}}.
\newblock Augmenting language models with long-term memory.
\newblock \emph{Advances in Neural Information Processing Systems}, 36.

\bibitem[{Wei et~al.(2022)Wei, Tay, Bommasani, Raffel, Zoph, Borgeaud, Yogatama, Bosma, Zhou, Metzler et~al.}]{Wei2022EmergentAO}
Jason Wei, Yi~Tay, Rishi Bommasani, Colin Raffel, Barret Zoph, Sebastian Borgeaud, Dani Yogatama, Maarten Bosma, Denny Zhou, Donald Metzler, et~al. 2022.
\newblock Emergent abilities of large language models.
\newblock \emph{Transactions on Machine Learning Research}.

\bibitem[{Yu et~al.(2024)Yu, Zhang, Hu, Storks, and Chai}]{yu2024eliciting}
Keunwoo Yu, Zheyuan Zhang, Fengyuan Hu, Shane Storks, and Joyce Chai. 2024.
\newblock Eliciting in-context learning in vision-language models for videos through curated data distributional properties.
\newblock In \emph{Proceedings of the 2024 Conference on Empirical Methods in Natural Language Processing}, pages 20416--20431.

\bibitem[{Zhang et~al.(2024{\natexlab{a}})Zhang, Lin, Yang, Wang, Li, Lin, Liu, and Wang}]{zhang2024mm}
Chaoyi Zhang, Kevin Lin, Zhengyuan Yang, Jianfeng Wang, Linjie Li, Chung-Ching Lin, Zicheng Liu, and Lijuan Wang. 2024{\natexlab{a}}.
\newblock Mm-narrator: Narrating long-form videos with multimodal in-context learning.
\newblock In \emph{Proceedings of the IEEE/CVF Conference on Computer Vision and Pattern Recognition}, pages 13647--13657.

\bibitem[{Zhang et~al.(2023)Zhang, Li, and Bing}]{zhang2023video}
Hang Zhang, Xin Li, and Lidong Bing. 2023.
\newblock \href {https://doi.org/10.18653/v1/2023.emnlp-demo.49} {Video-{LL}a{MA}: An instruction-tuned audio-visual language model for video understanding}.
\newblock In \emph{Proceedings of the 2023 Conference on Empirical Methods in Natural Language Processing: System Demonstrations}, pages 543--553, Singapore. Association for Computational Linguistics.

\bibitem[{Zhang et~al.(2022)Zhang, Roller, Goyal, Artetxe, Chen, Chen, Dewan, Diab, Li, Lin et~al.}]{zhang2022opt}
Susan Zhang, Stephen Roller, Naman Goyal, Mikel Artetxe, Moya Chen, Shuohui Chen, Christopher Dewan, Mona Diab, Xian Li, Xi~Victoria Lin, et~al. 2022.
\newblock Opt: Open pre-trained transformer language models.
\newblock \emph{arXiv preprint arXiv:2205.01068}.

\bibitem[{Zhang et~al.(2024{\natexlab{b}})Zhang, Bo, Ma, Li, Chen, Dai, Zhu, Dong, and Wen}]{zhang2024survey}
Zeyu Zhang, Xiaohe Bo, Chen Ma, Rui Li, Xu~Chen, Quanyu Dai, Jieming Zhu, Zhenhua Dong, and Ji-Rong Wen. 2024{\natexlab{b}}.
\newblock A survey on the memory mechanism of large language model based agents.
\newblock \emph{arXiv preprint arXiv:2404.13501}.

\bibitem[{Zheng et~al.(2023)Zheng, Chiang, Sheng, Zhuang, Wu, Zhuang, Lin, Li, Li, Xing et~al.}]{zheng2023judging}
Lianmin Zheng, Wei-Lin Chiang, Ying Sheng, Siyuan Zhuang, Zhanghao Wu, Yonghao Zhuang, Zi~Lin, Zhuohan Li, Dacheng Li, Eric Xing, et~al. 2023.
\newblock Judging llm-as-a-judge with mt-bench and chatbot arena.
\newblock \emph{Advances in Neural Information Processing Systems}, 36:46595--46623.

\bibitem[{Zhu et~al.()Zhu, Chen, Shen, Li, and Elhoseiny}]{zhuminigpt}
Deyao Zhu, Jun Chen, Xiaoqian Shen, Xiang Li, and Mohamed Elhoseiny.
\newblock Minigpt-4: Enhancing vision-language understanding with advanced large language models.
\newblock In \emph{The Twelfth International Conference on Learning Representations}.

\end{thebibliography}

\newpage
\appendix

\section{Notation table}
We summarize the notations used in the paper in Tab.~\ref{tab:notations}. 

\begin{table}[h!]
\centering
\begin{tabular}{ll}
\toprule
\textbf{Notation} & \textbf{Description} \\
\midrule
$\mathcal{E}$ & A video episode \\
$\mathcal{D}$ & Persistent memory \\
$\epmem$ & Memory (overall) \\
$\longmem$ & Long-term memory \\
$\shortmem$ & Short-term memory \\
$e_i$ & $i$-th event \\
$e_n$ & the current query event \\
$\narration{}{}$ & Narration of an event \\
$\gennarration{}{}$ & Generated narrations of an event \\
$(e, t)$ &  A pair of event and narration\\
$N_l$ & The size of long-term memory \\
$N_s$ & The size of short-term memory \\
$c$ & Semantic cluster of sampled narrations \\
$\LLM(\cdot)$ & generation of Large Language Model \\
\bottomrule
\end{tabular}
\caption{Notations used in the paper.}
\label{tab:notations}
\end{table}

\section{Related Works}
\subsection{Temporal Understanding in Videos}
Understanding temporal ques in videos is challenging and MLLM-based methods draw great attention. A wide range of work~\citep{maaz-etal-2024-video, qian2025streaming,zhang2024mm,he2024ma} leverages the reasoning ability of MLLMs and presents great performance in understanding holistic videos.

\subsection{Memory as Contexts in LLMs} Implementing memories has been a big topic. Due to the emergent abilities of LLMs~\citep{Wei2022EmergentAO} to understand and leverage context, using contexts as memory has been one mainstream approach for NLP tasks\citep{fountas2024human} and also multimodal tasks\citep{fan2024videoagent,chen2024camml}.

\section{Extension of Experimental Study}
\subsection{Model Implementation}\label{sec:more-on-model}
\paragraph{Similarity-based Retrieval} We retrieve relevant events from \longmem based on their semantic similarity with the current query event. We use the features in the output layer of the CLIP visual encoder and calculate their cosine similarity with the query event.

\paragraph{Recency-based Retrieval} We collect the events from \shortmem based on recency, namely the most $N_s$ recent events will be sampled from the current episode \video from the current event \event{}{n}.

\subsection{Training}\label{sec:training}
We elaborate our training algorithm as in Alg.~\ref{alg:training}. For training, we used 16 shots of memory events: 8 from long-term memory and 8 from short-term memory in the same episode. We use the ground-truth annotations in the training to adapt the MLLMs to learn the temporal and contextual knowledge from the memory.

We adopt the next token prediction loss. For training, we utilize 4 Nvidia A100 for 75 hours on the full training set of Ego4D with 5 epochs. The learning rate is set to 1e-5 and the batch size to 8. For streaming evaluation, we adopt 1 Nvidia A40 for 4 hours.

\begin{algorithm}
\caption{Training}
\label{alg:training}
\KwIn{Current query event \event{}{n}, \\Current episode \video, \\ Persistent memory \data}
\KwOut{Prediction of narration \narration{}{n}}
Collect short-term memory from the current episode \video \\
\For{$j = 1$ to $N_s$}{
    $\shortmem \gets (\event{}{n-j}, \narration{}{n-j}) \in \video$\;
}
Collect long-term memory from \data \\
\For{$k = 1$ to $N_l$}{
    $\longmem \gets (\event{l}{k}, \narration{}{k}) \in \data$\;
}
$\epmem \gets \longmem \cup \shortmem$ \;
$\narration{}{n} \gets \LLM(\epmem, \event{}{n})$ \;
\Return \narration{}{n}\;
\end{algorithm}

\subsection{\cameo Implementation}\label{sec:cameo-more}
\begin{figure*}
    \centering
    \begin{subfigure}{0.49\linewidth}
        \centering
        \includegraphics[width=\linewidth]{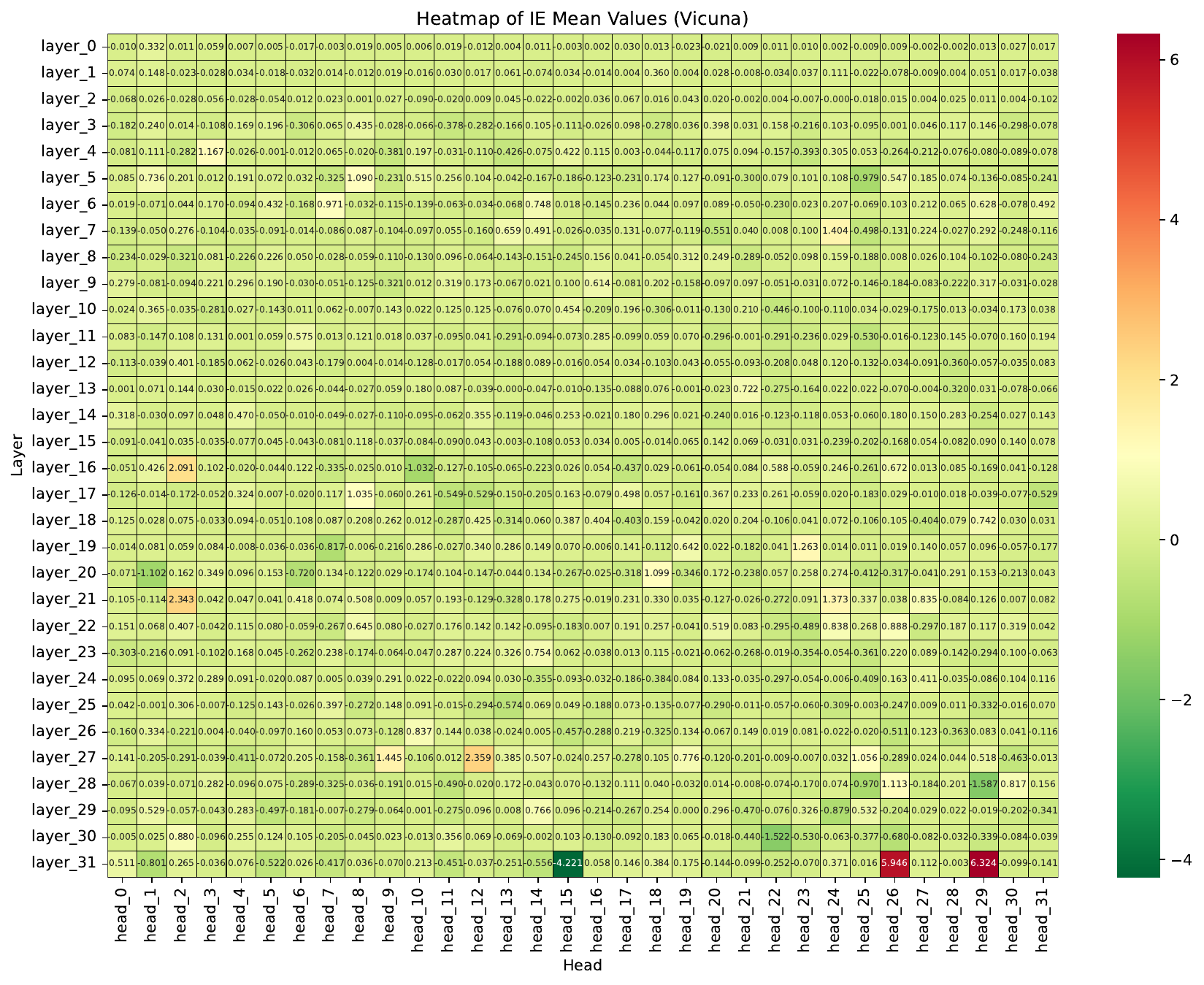}
        \caption{Probing result of \vicuna in different layers and heads}
        \label{fig:vicuna-prob}
    \end{subfigure}
    \hfill
    \begin{subfigure}{0.49\linewidth}
        \centering
        \includegraphics[width=\linewidth]{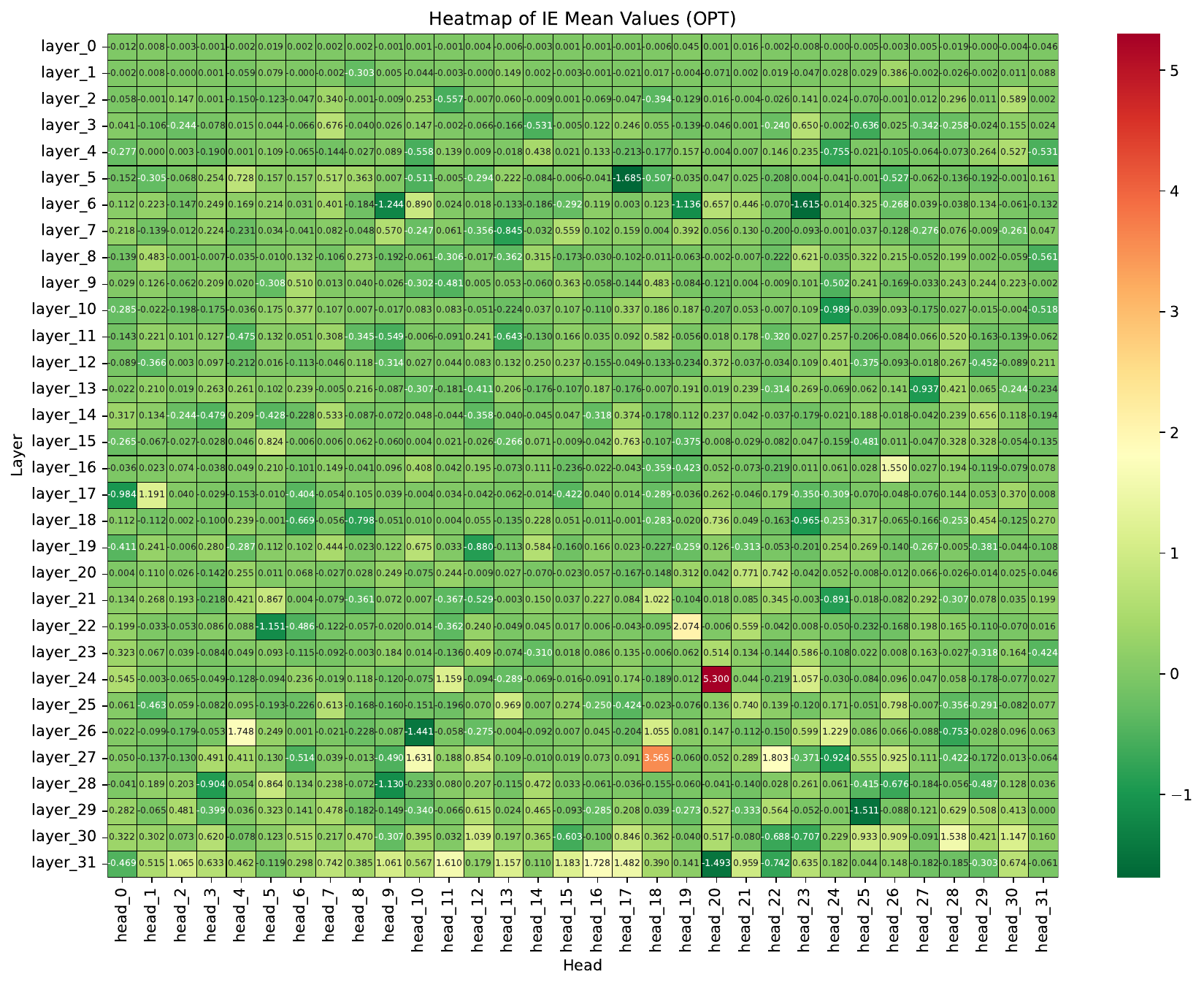}
        \caption{Probing result of \opt in different layers and heads}
        \label{fig:opt-prob}
    \end{subfigure}
    \caption{Comparison of probing results for Vicuna-7b and OPT-2.7b across different layers and heads.}
    \label{fig:probing-comparison}
\end{figure*}

\paragraph{Probing Confabulation-prone Attention Heads}
We selected 9 episodes consisting of 189 events from the validation split of Ego4D. For each event, we compose the memory set \epmem with 8 long-term memories and 8 ground-truth short-term memories and randomly replace one narration of a short-term event with a wrong narration. Then we ablate each attention head in each layer to calculate the influential effect. As shown in Fig.~\ref{fig:probing-comparison}, we locate the top-$k$ confabulation-prone attention heads with the largest IE.

In the end, we pick up top-96 confabulation-prone attention heads and use a temperature of 0.6 for \vicuna; top-32 confabulation-prone attention heads and use a temperature of 0.8 for \opt.

\paragraph{Attention Modification in CAMEO}
Considering the memory as contexts $\mathcal{C}=\{\event{l}{1}, \narration{l}{1}, \dots,\event{l}{1}, \narration{l}{N_l}, \event{s}{N_l}, \gennarration{s}{1}, \dots,\event{s}{N_s}, \gennarration{s}{N_s}\}$, each generated narration \gennarration{}{} has its semantic entropy score $se(\gennarration{s}{})$ and its modification weight $w(\gennarration{s}{})$. For other contextual inputs like $\event{}{}$ and $\narration{l}{}$, the modification weight is set to $1$. The resulting modification weight is $\mathbf{w}$ of which $k$-th token has the weight,
\begin{equation}
\mathbf{w}_k =
\begin{cases} 
w(\gennarration{s}{}) & \text{if } token_k \text{ belongs to } \gennarration{s}{}, \\
1 & \text{otherwise}
\end{cases}
\end{equation}

For an arbitrary attention head $h$ in transformers, we multiply the attention matrix $\mathbf{A}_h$ by the modification weight $\mathbf{w}$ elementwisely $\mathbf{A}_h \circ \mathbf{w}$.

\subsection{Extended Results}

\begin{figure*}[htbp]
    \centering
    \begin{minipage}{0.48\textwidth}
        \centering
        \begin{subfigure}{0.48\textwidth}
            \centering
            \includegraphics[width=\textwidth]{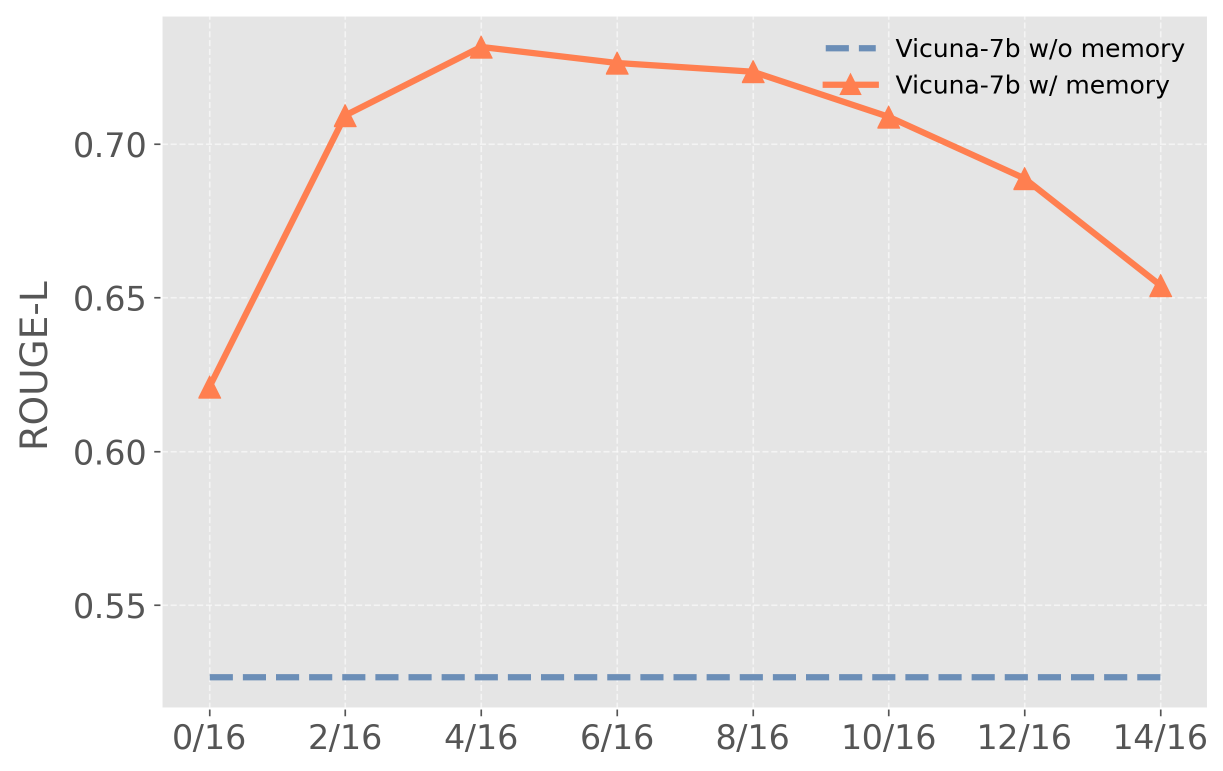}
            \caption{ROUGE-L (25\% short-term memory in training)}
            \label{fig:minigpt4_1_4_fmeasure}
        \end{subfigure}
        \hfill
        \begin{subfigure}{0.48\textwidth}
            \centering
            \includegraphics[width=\textwidth]{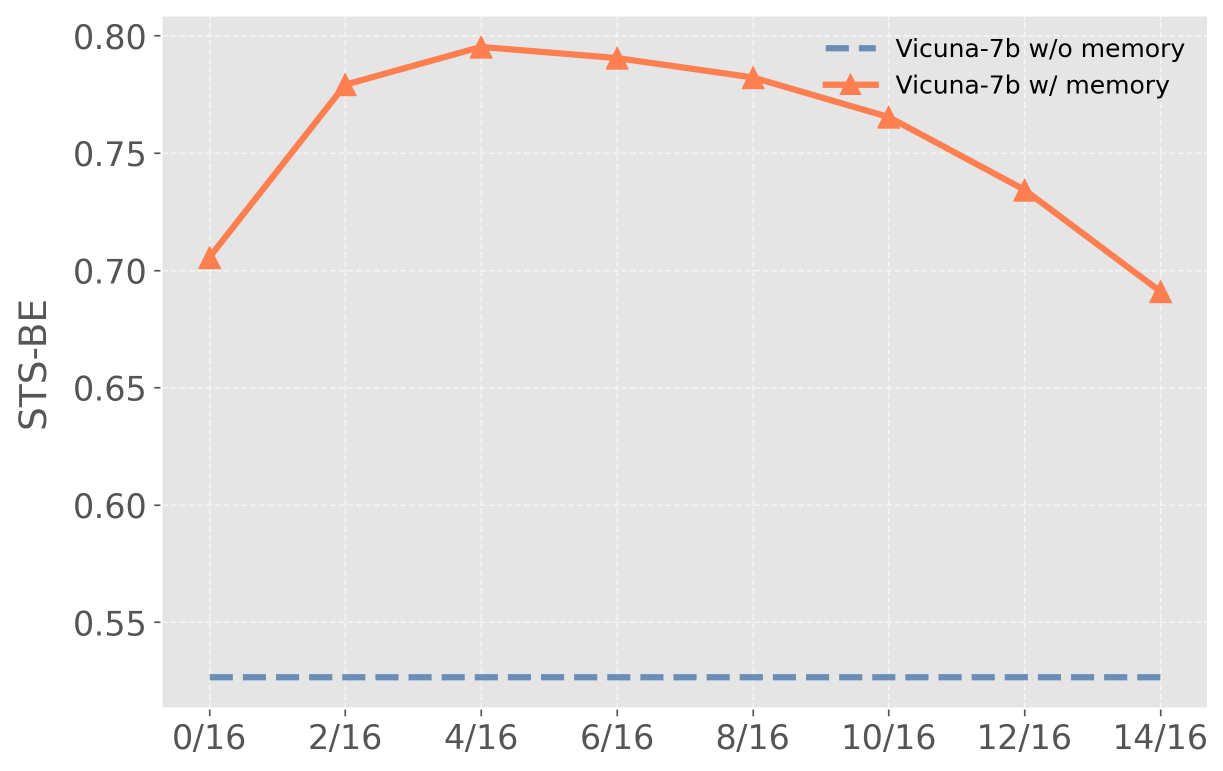}
            \caption{STS (25\% short-term memory in training)}
            \label{fig:minigpt4_1_4_sbs}
        \end{subfigure}
        
        \begin{subfigure}{0.48\textwidth}
            \centering
            \includegraphics[width=\textwidth]{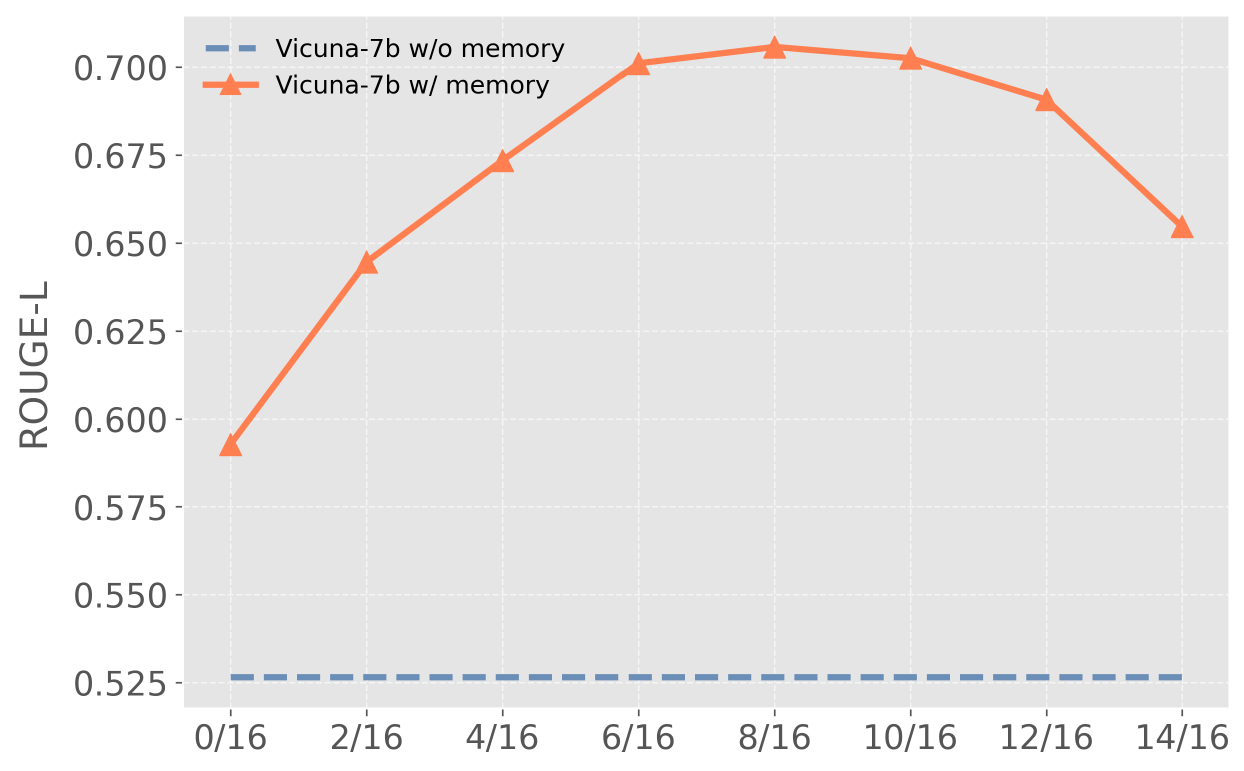}
            \caption{ROUGE-L (50\% short-term memory in training)}
            \label{fig:minigpt4_1_2_fmeasure}
        \end{subfigure}
        \hfill
        \begin{subfigure}{0.48\textwidth}
            \centering
            \includegraphics[width=\textwidth]{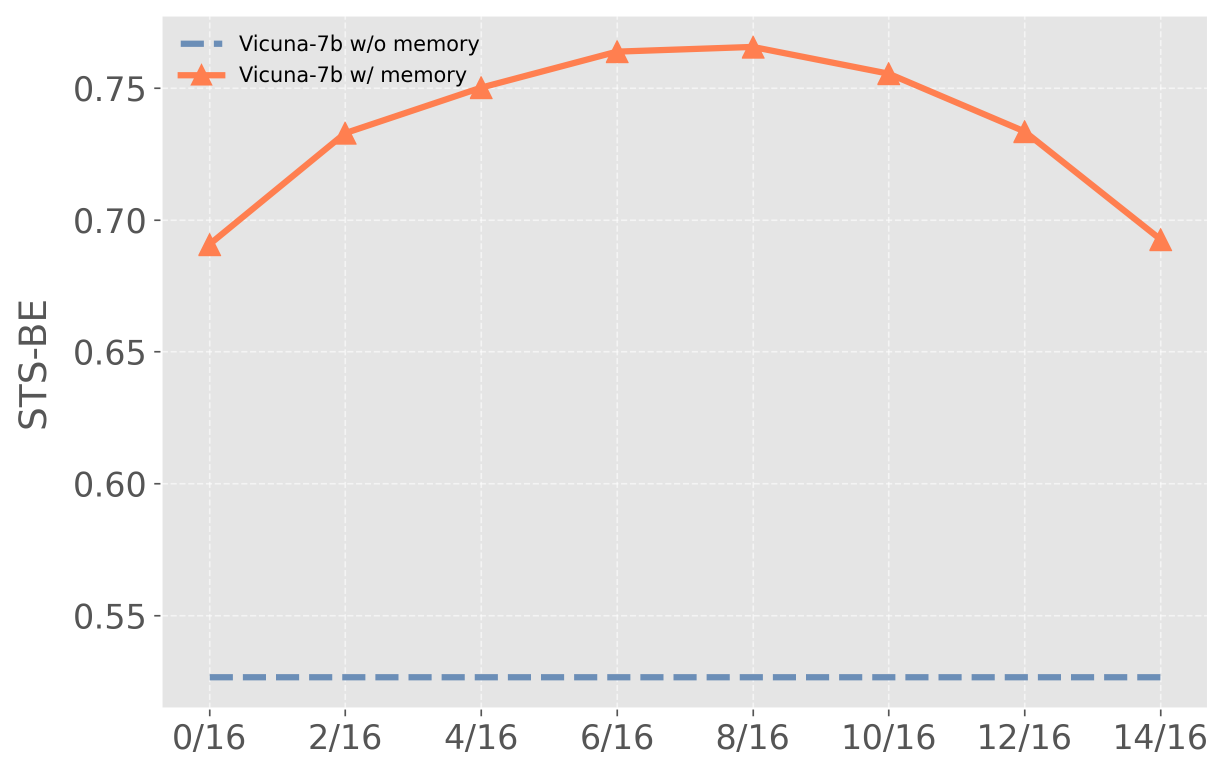}
            \caption{STS (50\% short-term memory in training)}
            \label{fig:minigpt4_1_2_sbs}
        \end{subfigure}
        
        \begin{subfigure}{0.48\textwidth}
            \centering
            \includegraphics[width=\textwidth]{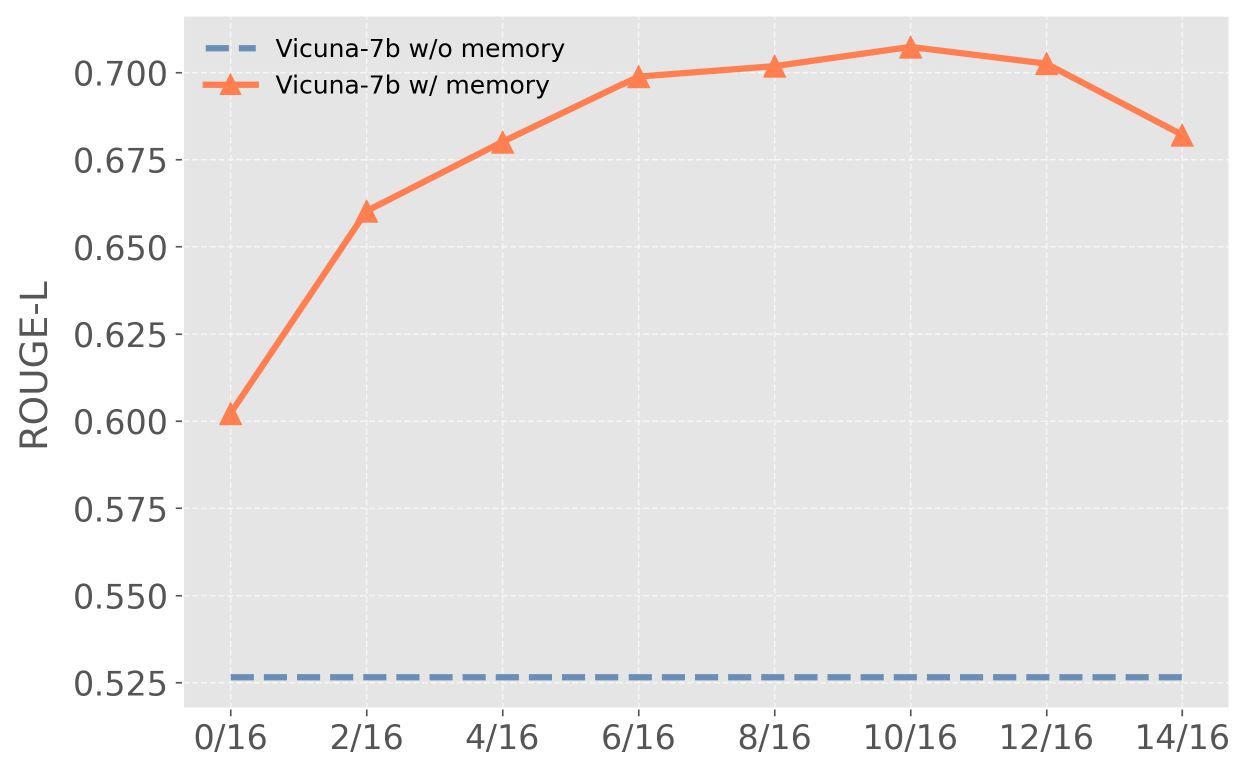}
            \caption{ROUGE-L (75\% short-term memory in training)}
            \label{fig:minigpt4_3_4_fmeasure}
        \end{subfigure}
        \hfill
        \begin{subfigure}{0.48\textwidth}
            \centering
            \includegraphics[width=\textwidth]{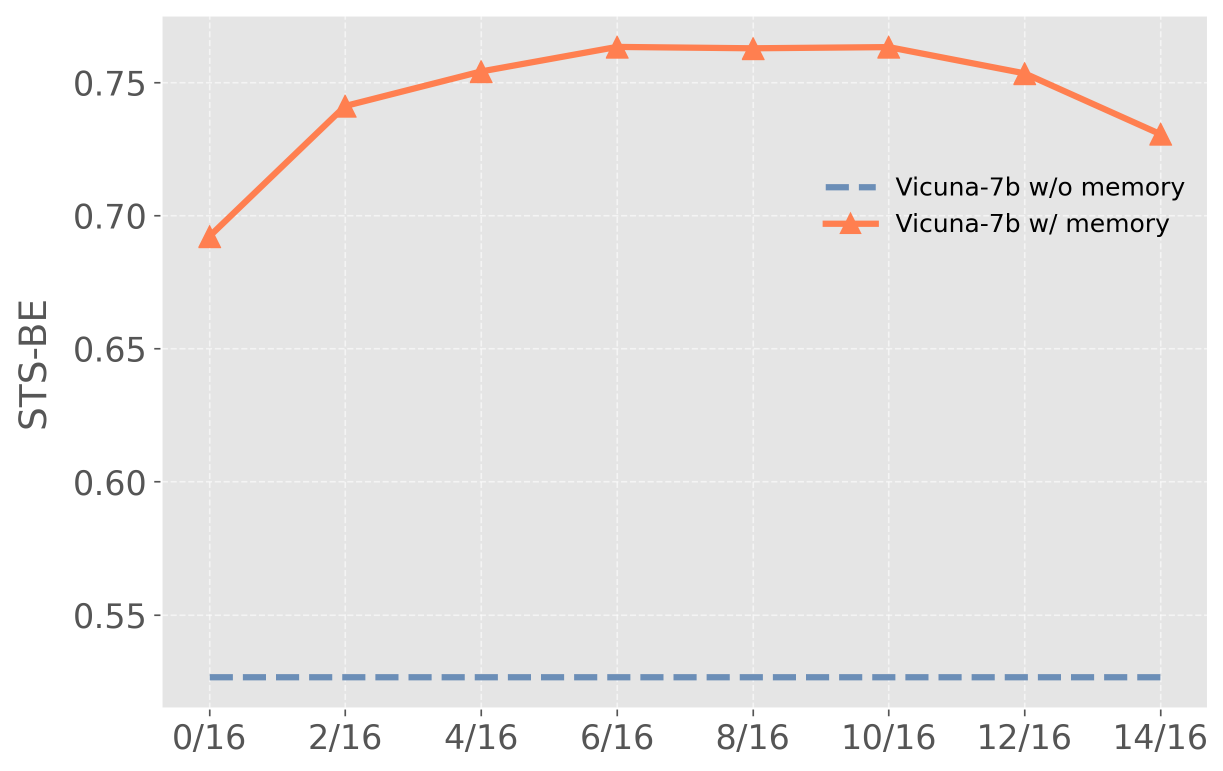}
            \caption{STS (75\% short-term memory in training)}
            \label{fig:minigpt4_3_4_sbs}
        \end{subfigure}
        \caption{Evaluation of Vicuna-7b with different training ratios.}
        \label{fig:minigpt4_eval}
    \end{minipage}
    \hfill
    \begin{minipage}{0.48\textwidth}
        \centering

        \begin{subfigure}[t]{0.48\textwidth}
            \centering
            
            \includegraphics[width=\textwidth]{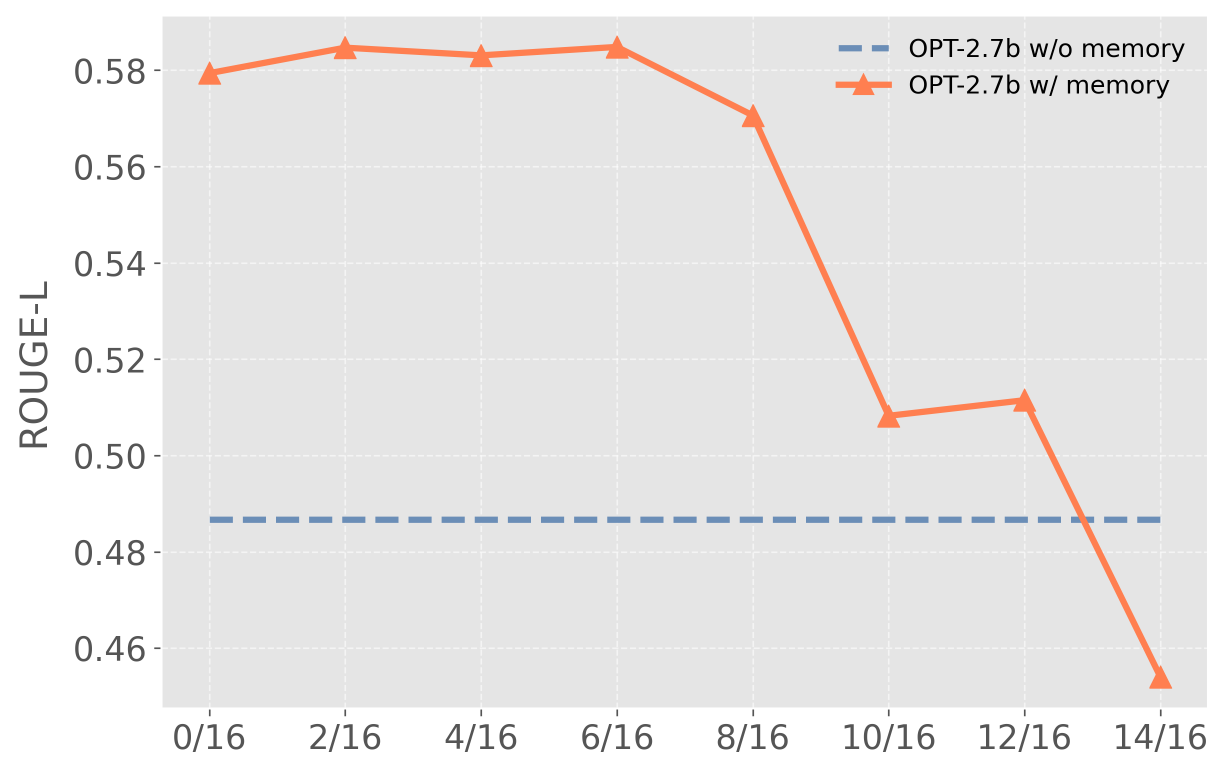}
            \caption{ROUGE-L (25\% short-term memory in training)}
            \label{fig:opt_1_4_fmeasure}
        \end{subfigure}
        \hfill
        \begin{subfigure}[t]{0.48\textwidth}
            \centering
            \includegraphics[width=\textwidth]{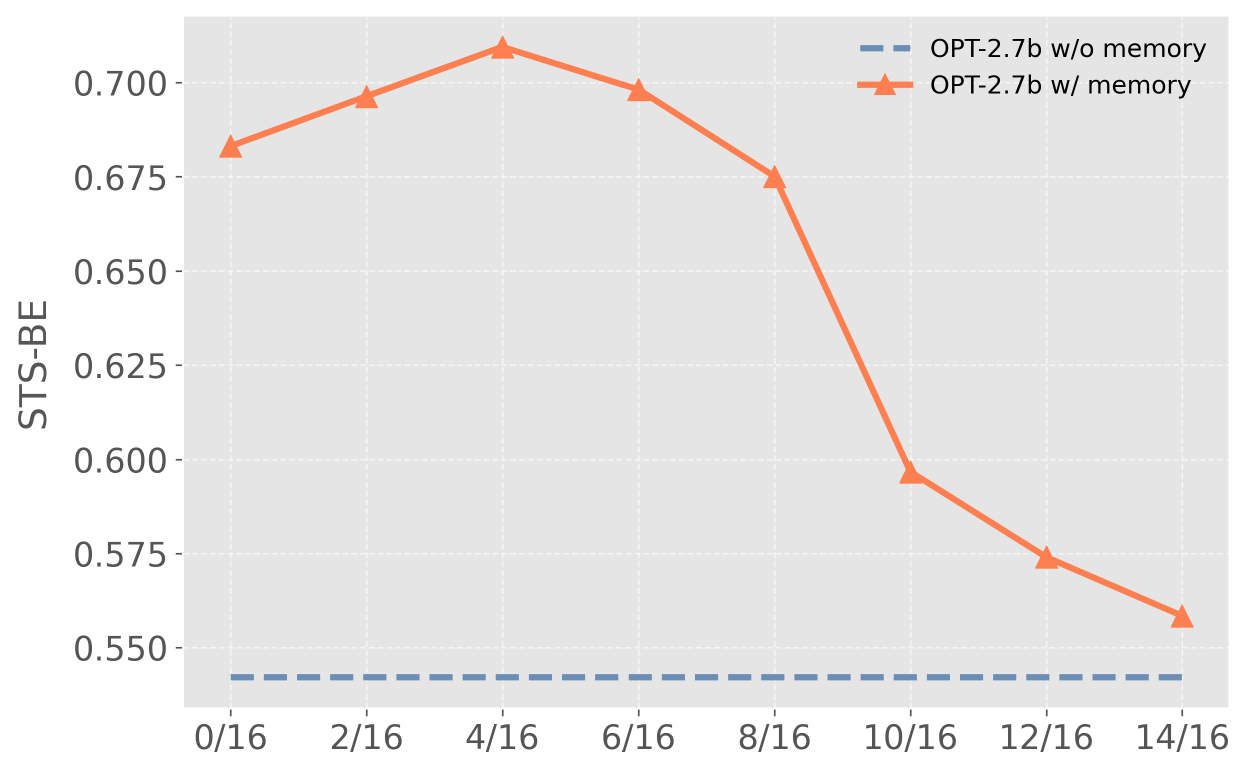}
            \caption{STS (25\% short-term memory in training)}
            \label{fig:opt_1_4_sbs}
        \end{subfigure}
        
        \begin{subfigure}{0.48\textwidth}
            \centering
            \includegraphics[width=\textwidth]{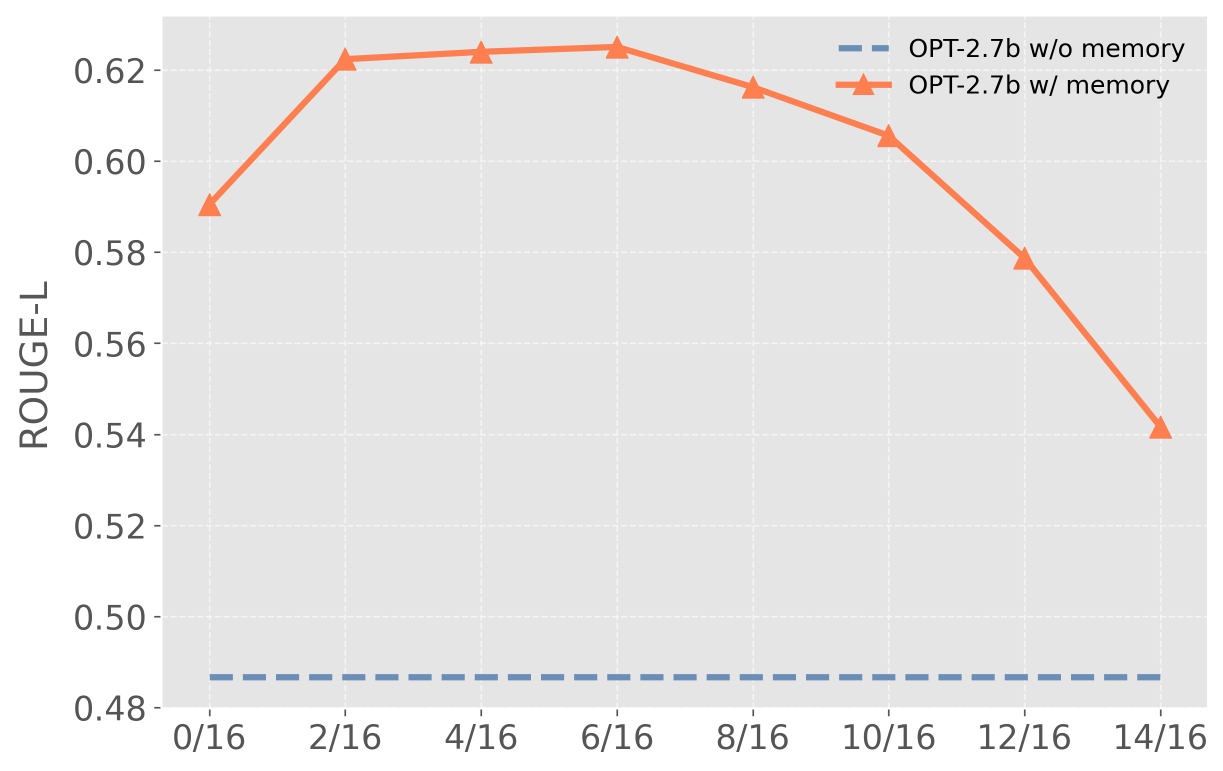}
            \caption{ROUGE-L (50\% short-term memory in training)}
            \label{fig:opt_1_2_fmeasure}
        \end{subfigure}
        \hfill
        \begin{subfigure}{0.48\textwidth}
            \centering
            \includegraphics[width=\textwidth]{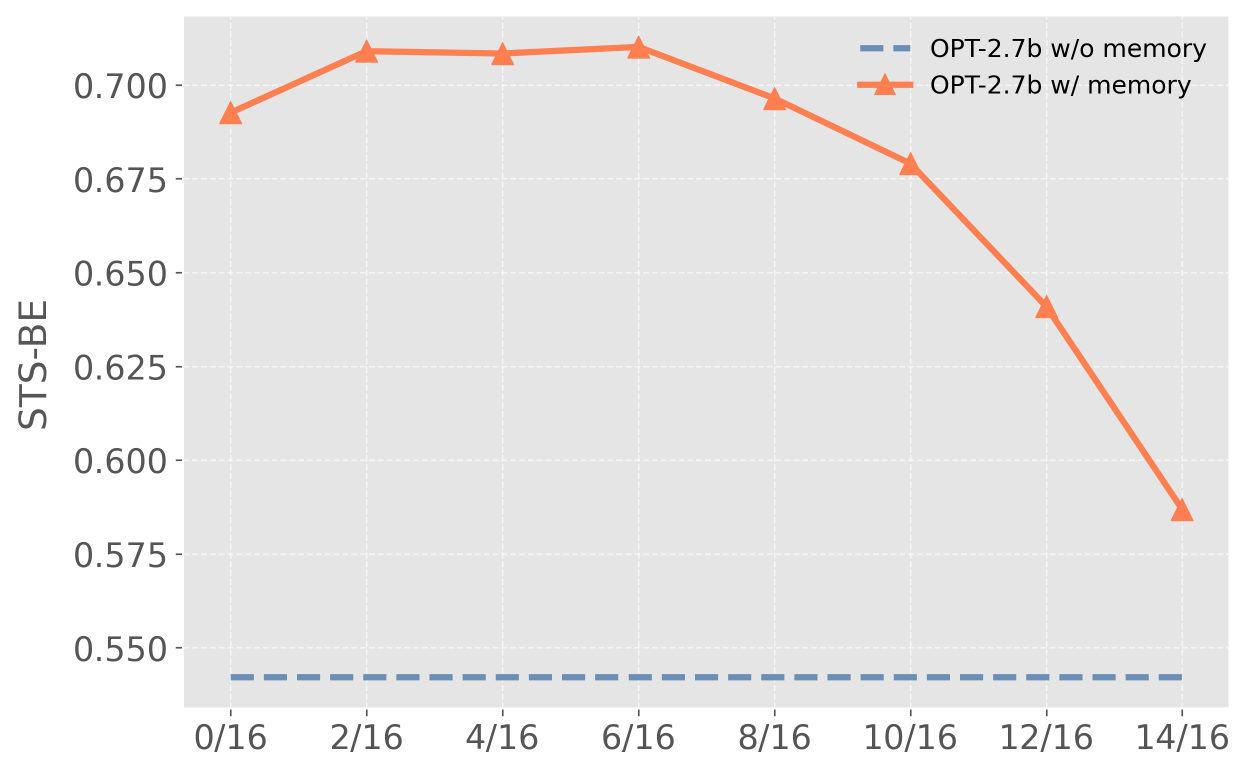}
            \caption{STS (50\% short-term memory in training)}
            \label{fig:opt_1_2_sbs}
        \end{subfigure}
        
        \begin{subfigure}{0.48\textwidth}
            \centering
            \includegraphics[width=\textwidth]{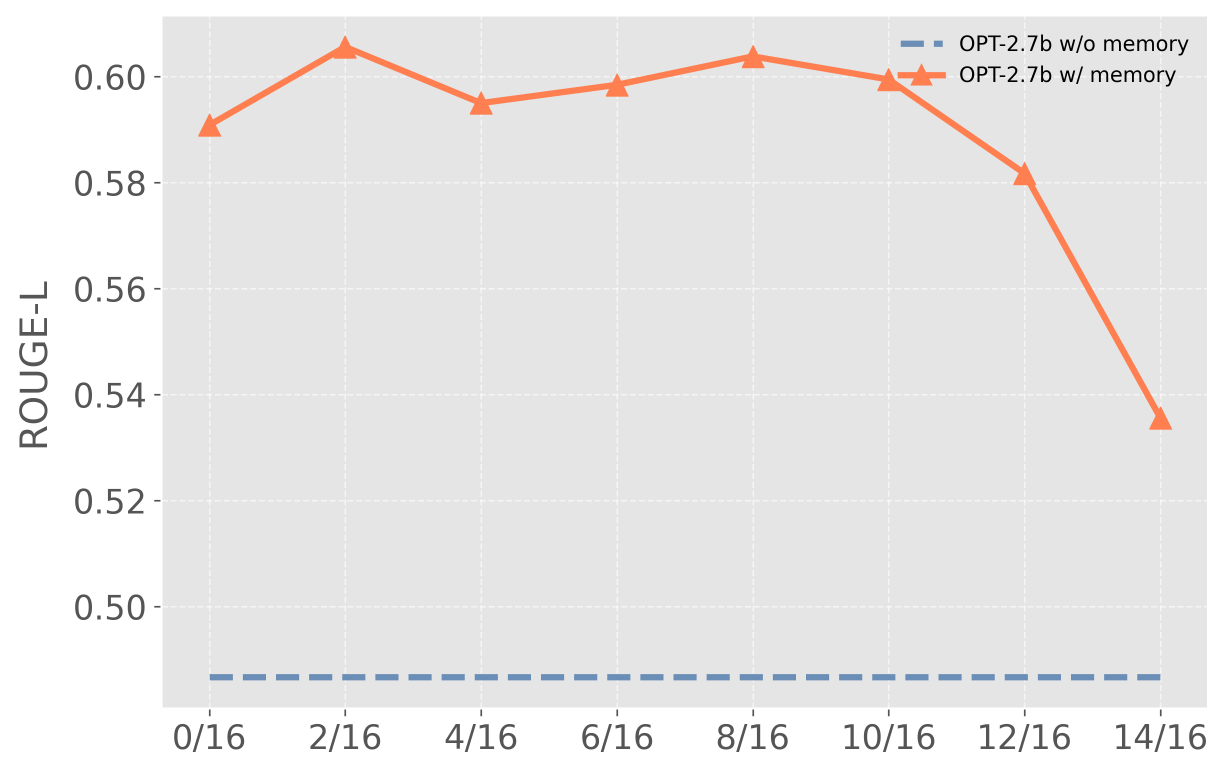}
            \caption{ROUGE-L (75\% short-term memory in training)}
            \label{fig:opt_3_4_fmeasure}
        \end{subfigure}
        \hfill
        \begin{subfigure}{0.48\textwidth}
            \centering
            \includegraphics[width=\textwidth]{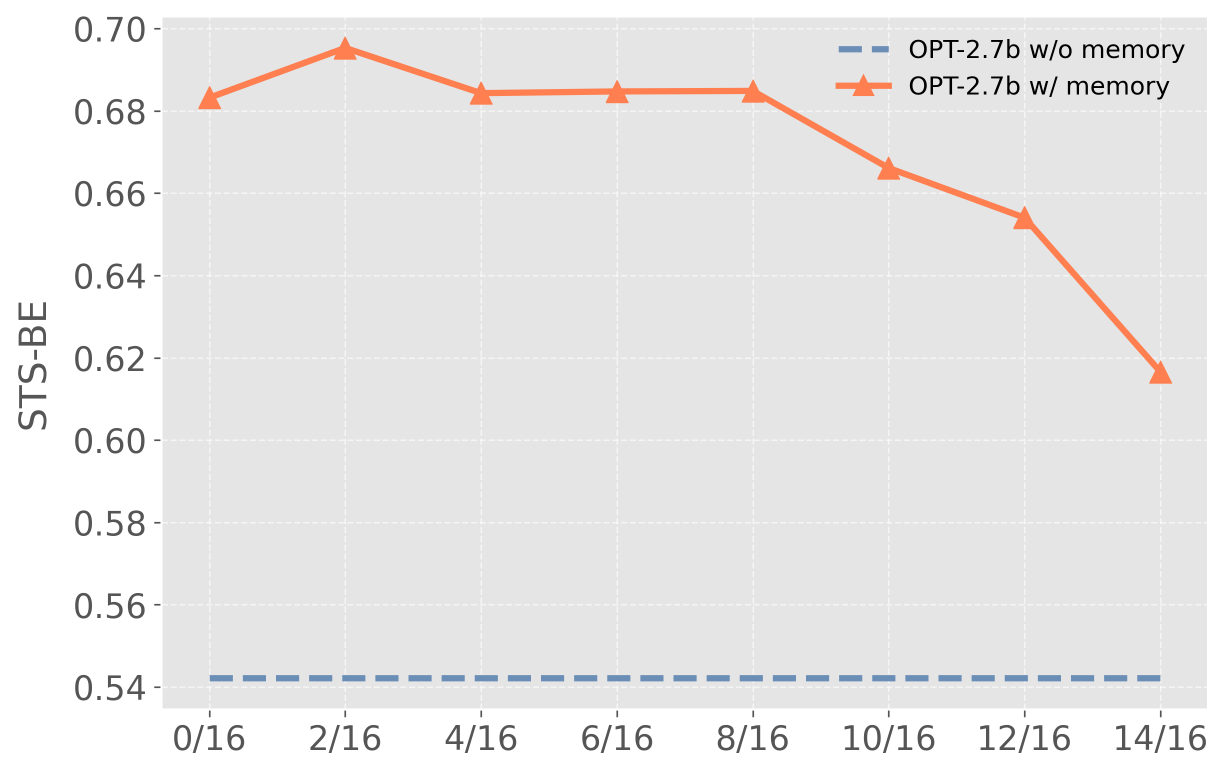}
            \caption{STS (75\% short-term memory in training)}
            \label{fig:opt_3_4_sbs}
        \end{subfigure}
        \caption{Evaluation of \opt with different ratios.}
        \label{fig:opt_eval}
    \end{minipage}
\end{figure*}

\paragraph{Different ratio of memory}
We explore different ratios of short-term memory \shortmem and long-term memory \longmem. We defined the total length of \epmem as $16$ and tried different ratios of \shortmem over a total length of \epmem.
As shown in Fig.~\ref{fig:minigpt4_eval}-\ref{fig:opt_eval}, we experiment with \vicuna and \opt under various ratios of memory in the training and evaluation stages (using ground-truth memory) respectively. With the memoryless baseline (0-shot) highlighted in the figures, we observe that a blend of short-term and long-term memory strikes a strong balance in event understanding performance. We also find that this trade-off is somewhat impacted by the ratio applied in the training phase.
Furthermore, all memory-augmented settings consistently outperform the memoryless baseline, thus reinforcing our finding that memory indeed aids event understanding.

\begin{figure*}[htbp]
    \centering
    \begin{subfigure}[b]{0.3\textwidth}
        \centering
        \includegraphics[width=\textwidth]{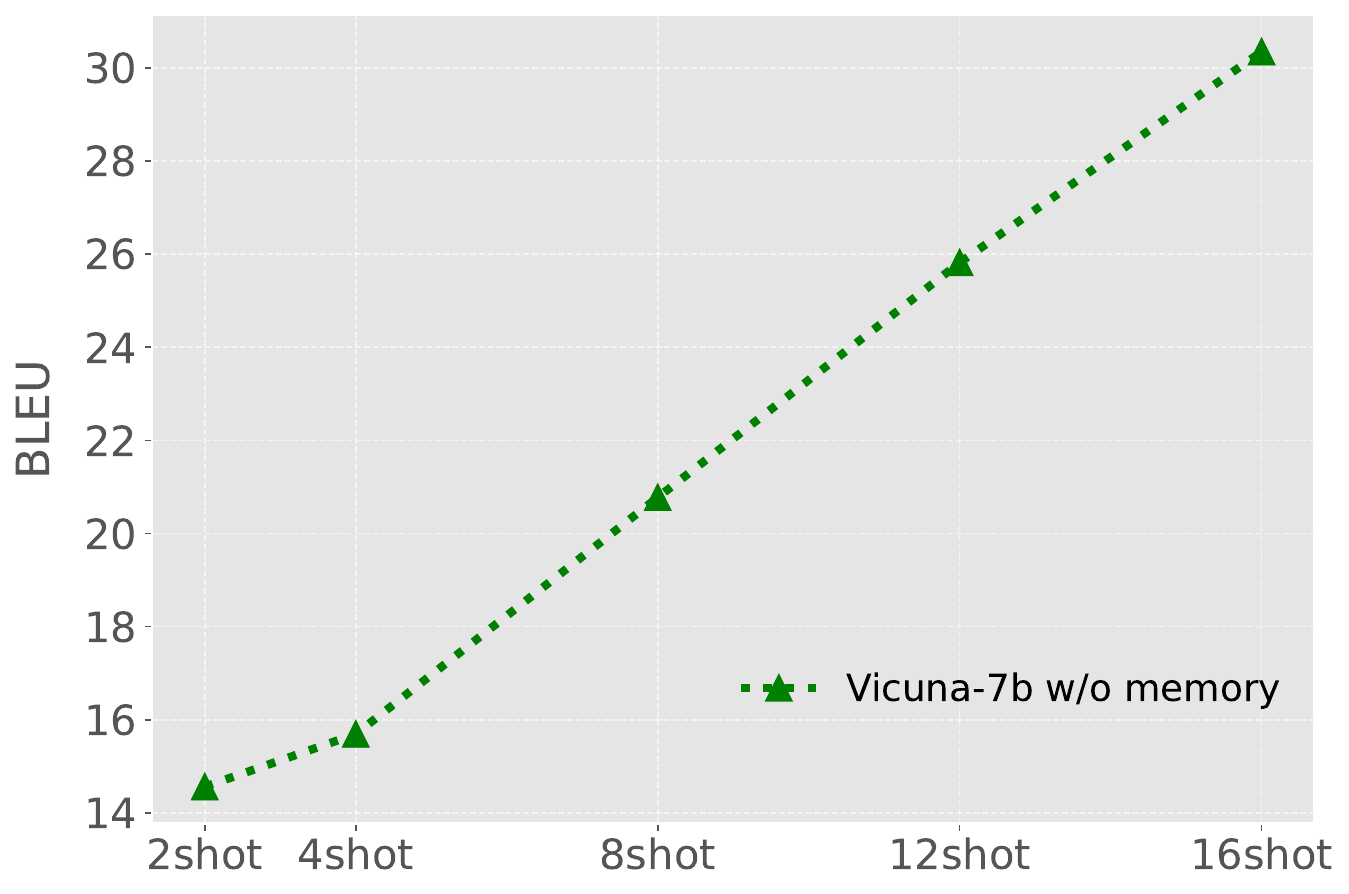}
        \caption{BLEU}
        \label{fig:mini-gt-BLEU}
    \end{subfigure}
    \hfill
    \begin{subfigure}[b]{0.3\textwidth}
        \centering
        \includegraphics[width=\textwidth]{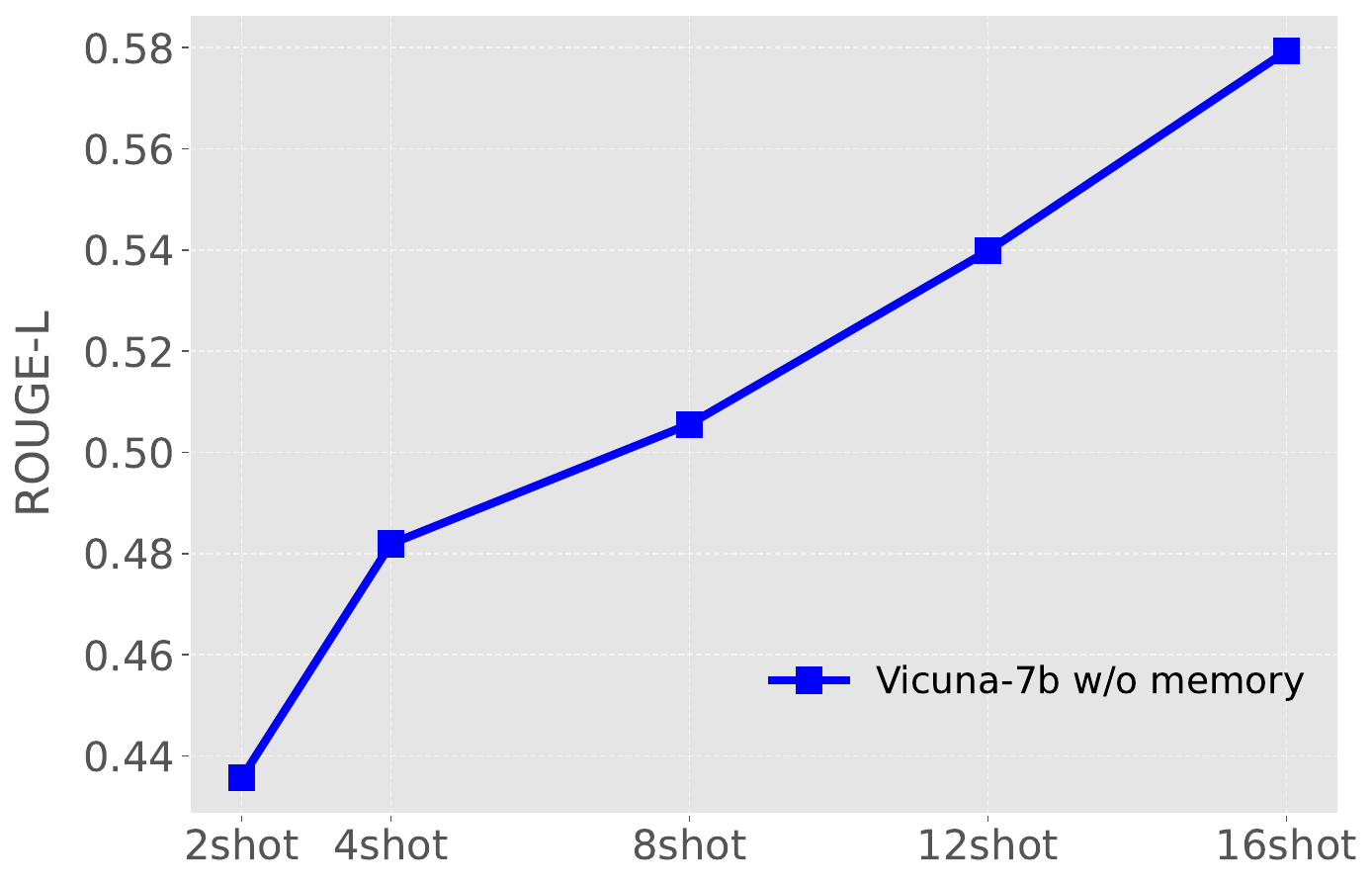}
        \caption{ROUGE-L}
        \label{fig:mini-gt-rouge}
    \end{subfigure}
    \hfill
    \begin{subfigure}[b]{0.3\textwidth}
        \centering
        \includegraphics[width=\textwidth]{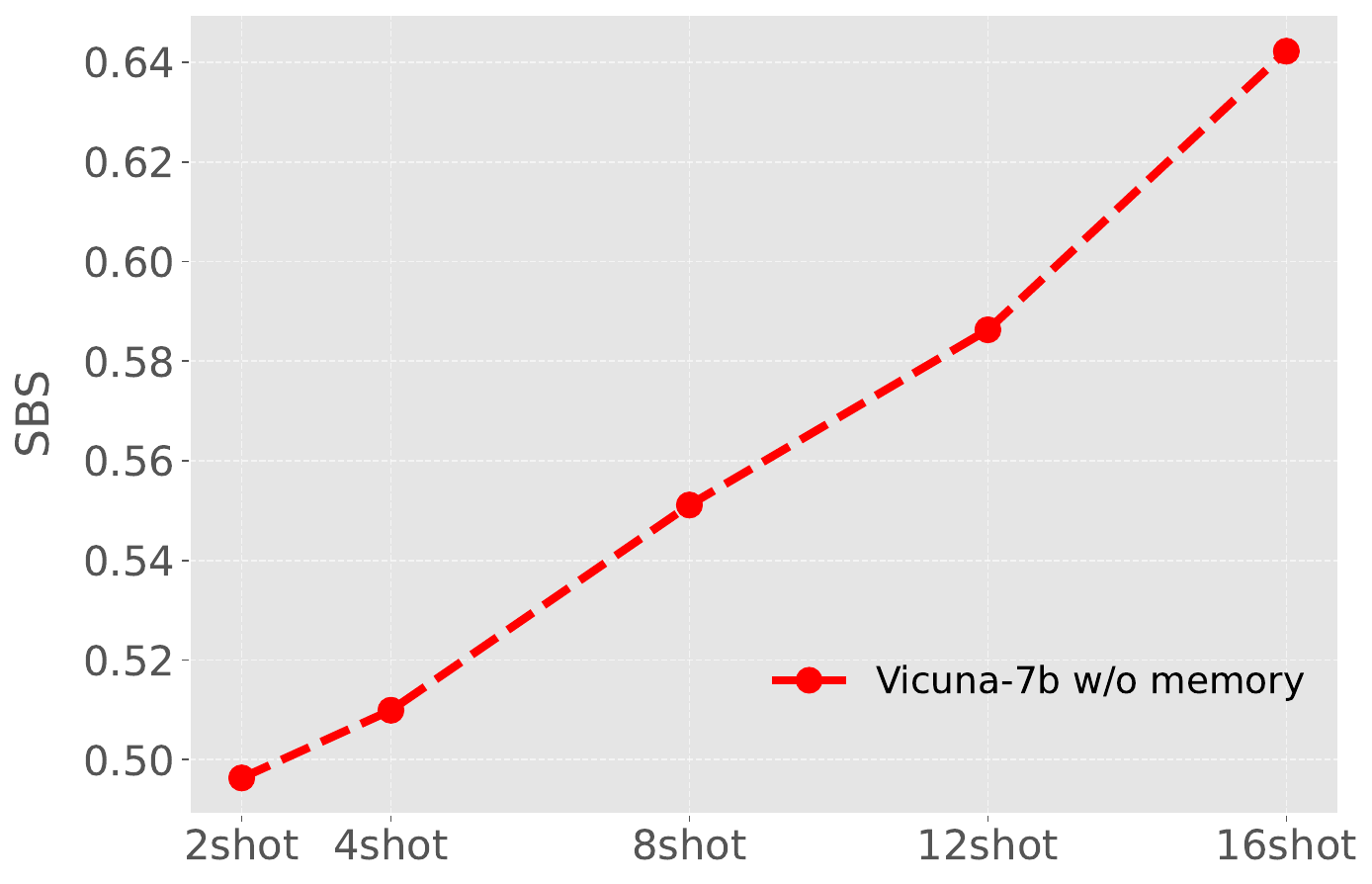}
        \caption{STS}
        \label{fig:mini-gt-sbs}
    \end{subfigure}
    \caption{Results of \vicuna with ground-truth (GT) memory of different shots. We employ $50\%$ short-term memory in all the experiments.}
    \label{fig:mini-gt}
\end{figure*}

\begin{figure*}[t]
    \centering
    \begin{subfigure}[b]{0.3\textwidth}
        \centering
        \includegraphics[width=\textwidth]{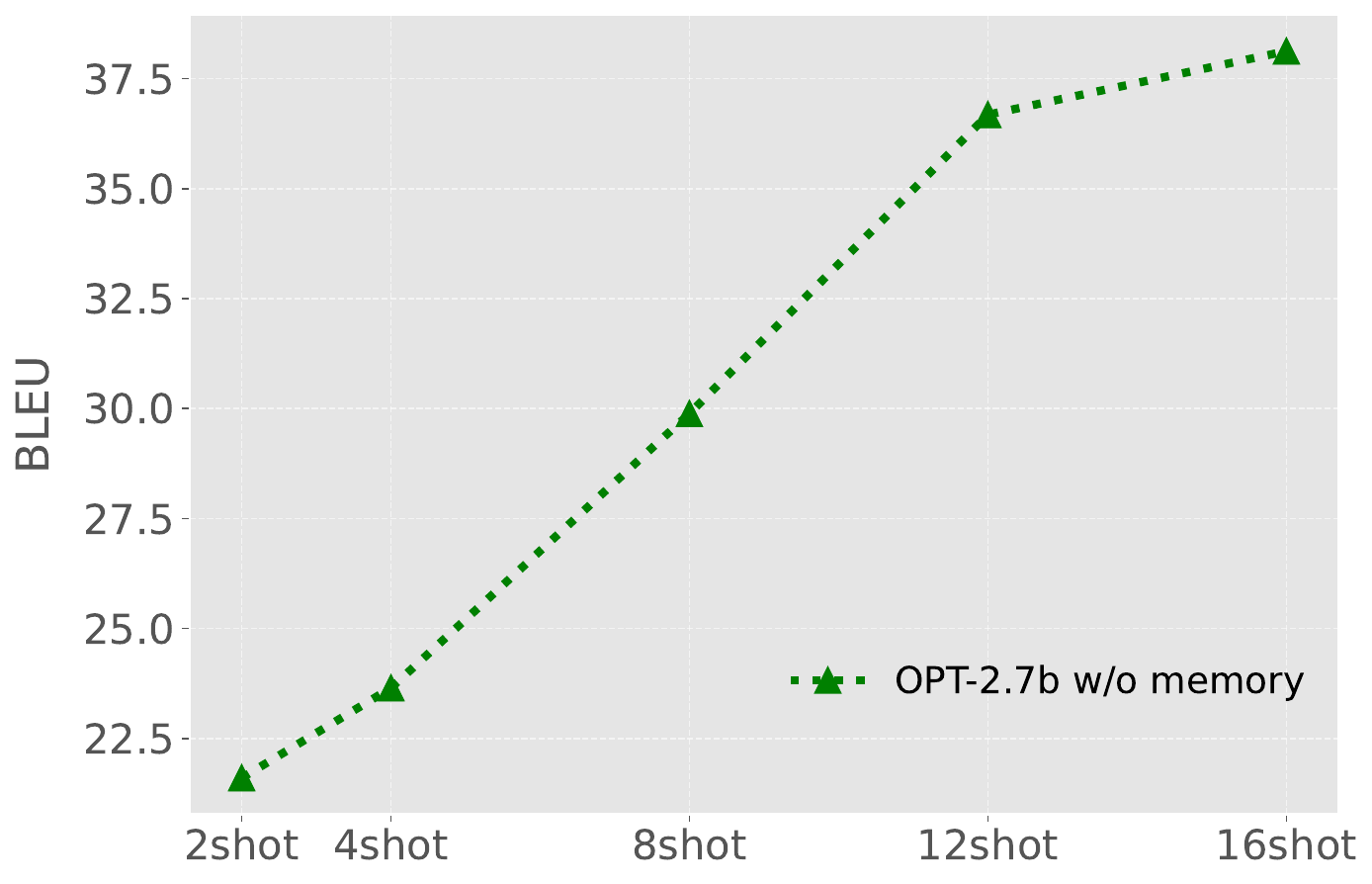}
        \caption{BLEU}
        \label{fig:mini-gt-BLEU}
    \end{subfigure}
    \hfill
    \begin{subfigure}[b]{0.3\textwidth}
        \centering
        \includegraphics[width=\textwidth]{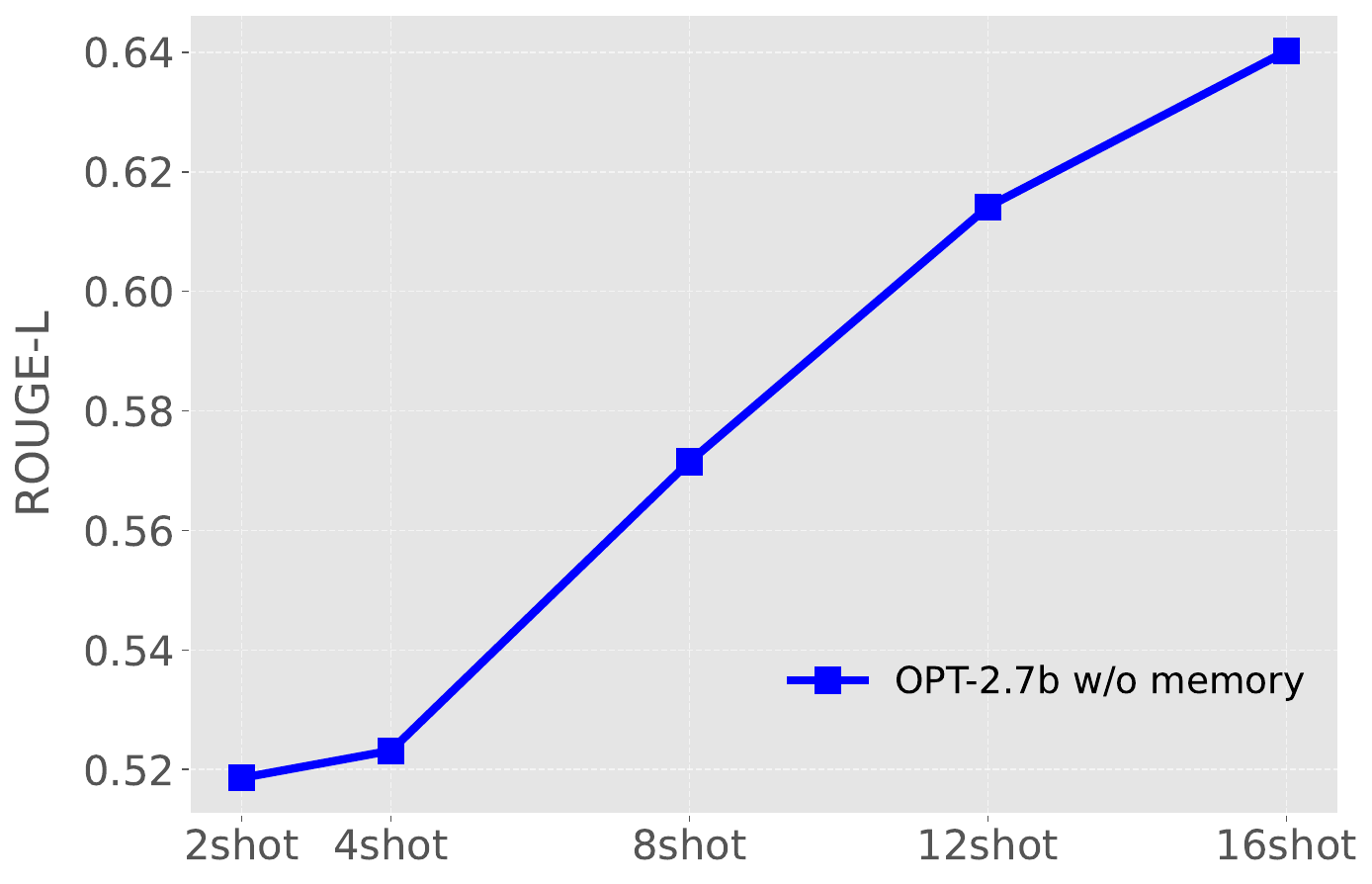}
        \caption{ROUGE-L}
        \label{fig:mini-gt-rouge}
    \end{subfigure}
    \hfill
    \begin{subfigure}[b]{0.3\textwidth}
        \centering
        \includegraphics[width=\textwidth]{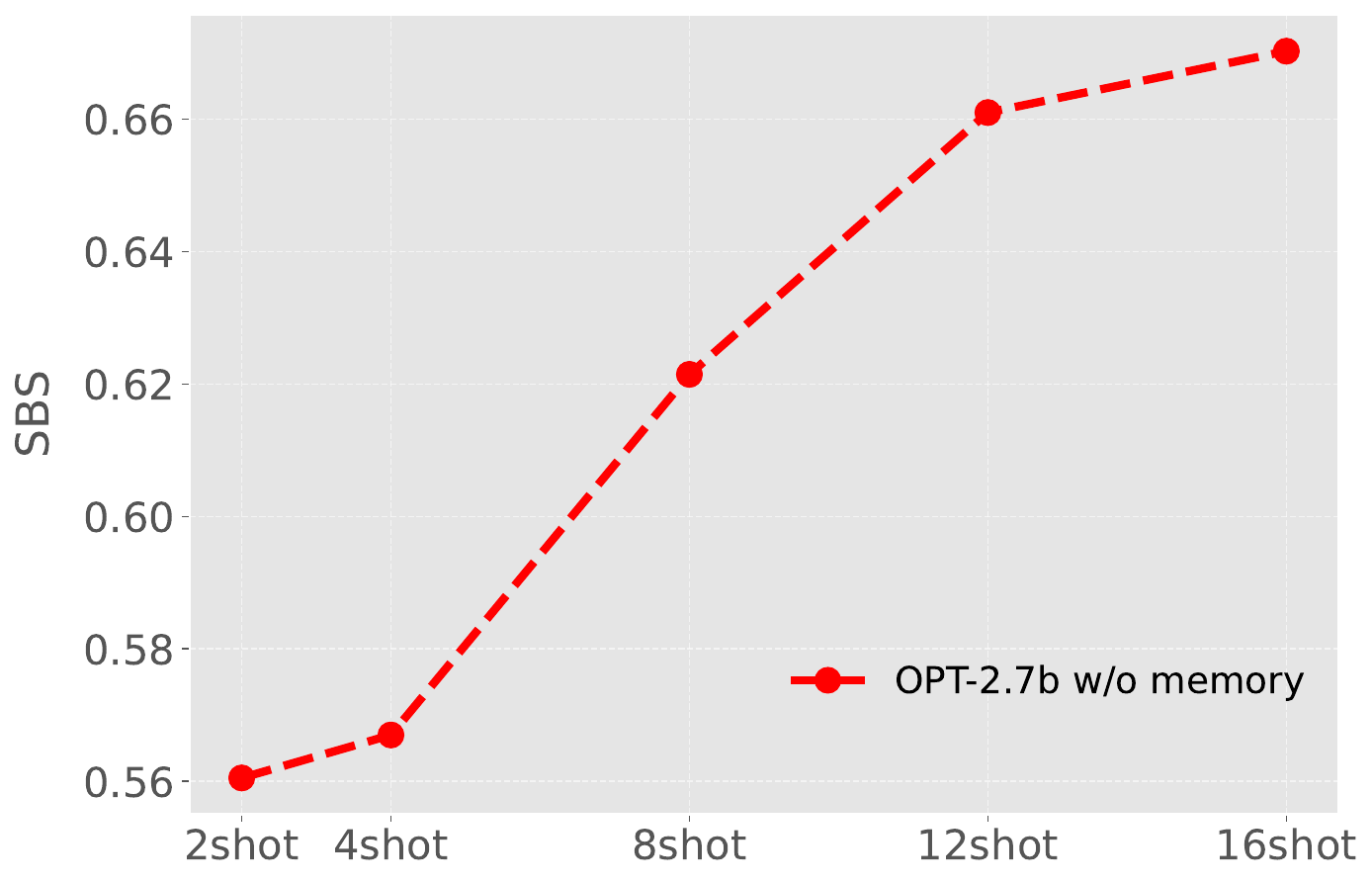}
        \caption{STS}
        \label{fig:mini-gt-sbs}
    \end{subfigure}
    \caption{Results of \vicuna with ground-truth (GT) memory of different shots.}
    \label{fig:opt-gt}
\end{figure*}

\paragraph{Different shots of memory}
We also study the performance impact of different shots. We use the same ratio $1/2$ where half of the memory is short-term memory as in Fig.~\ref{fig:mini-gt}-\ref{fig:opt-gt}.
We find that an increasing number of events in memory can increase the performance. This aligns with our emphasis over the impact of memory and the in-context learning ability in LLMs \citep{yu2024eliciting, brown2020language}.

\vspace{-0.2cm}

\subsection{Qualitative Studies}
\begin{figure*}
    \begin{subfigure}{\textwidth}
        \centering
        \includegraphics[width=0.8\textwidth]{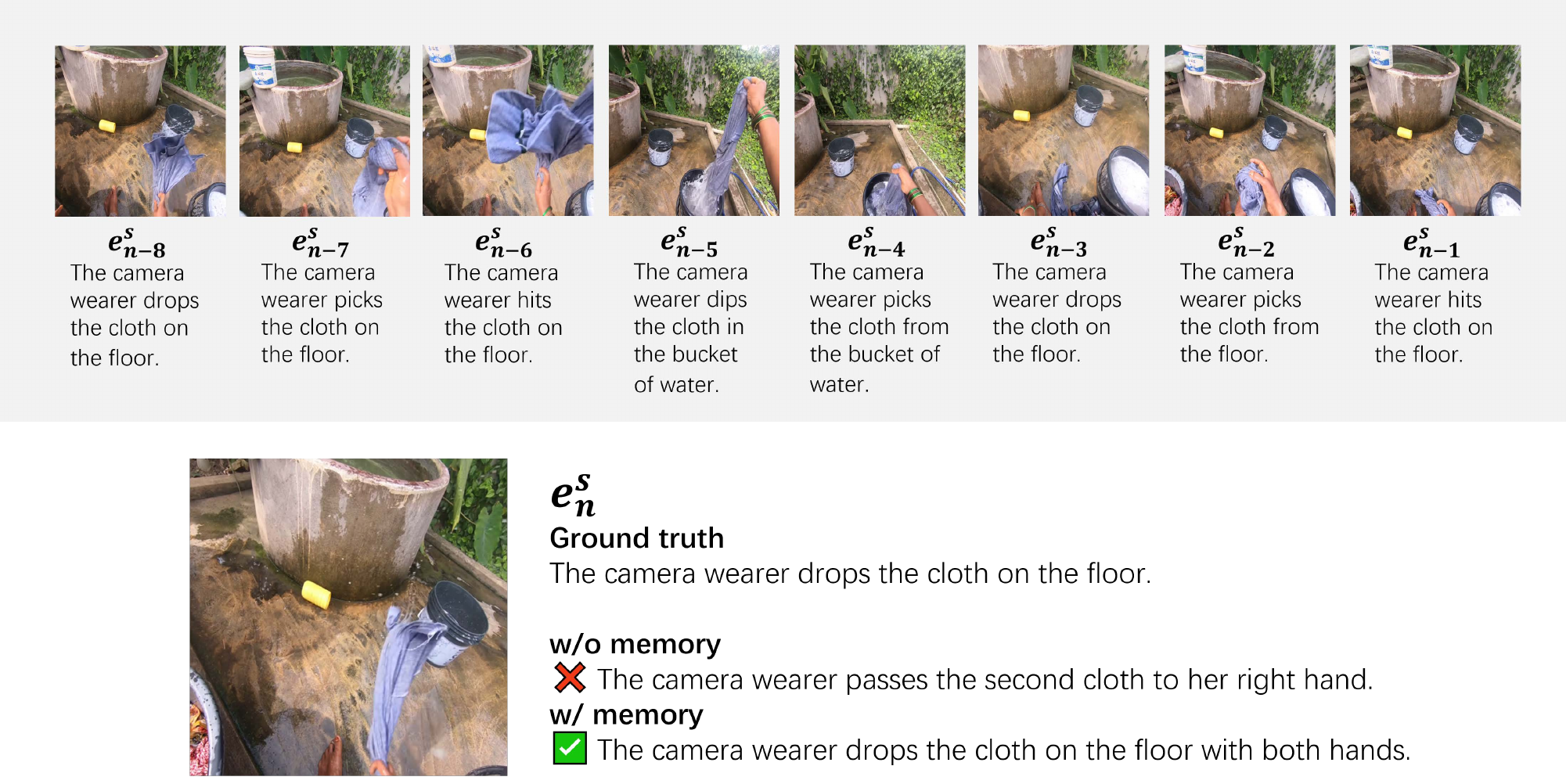}
        \caption{Example 1}
        \label{fig:ex1}
    \end{subfigure}

    \begin{subfigure}{\textwidth}
        \centering
        \includegraphics[width=0.8\textwidth]{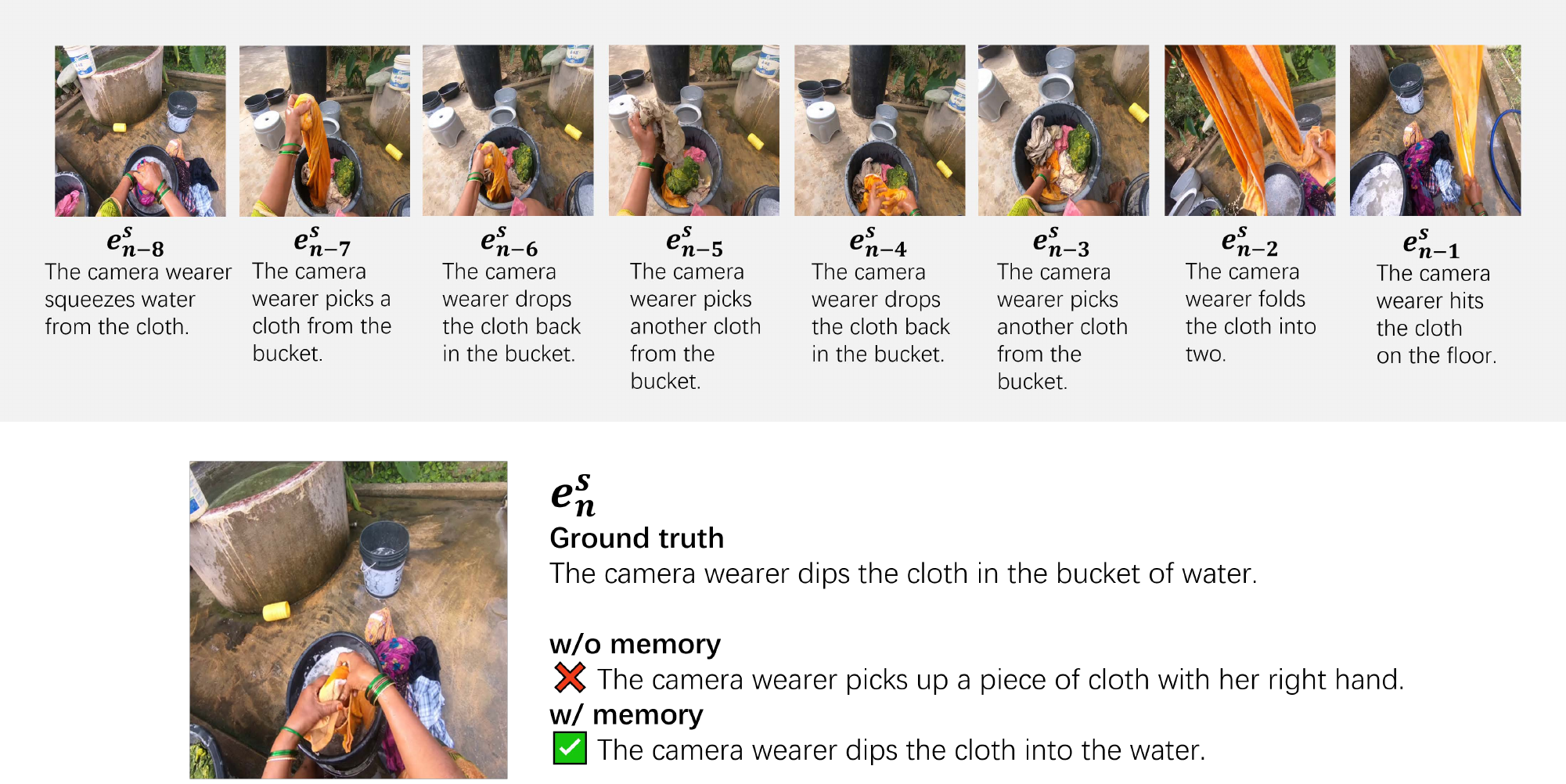}
        \caption{Example 2}
        \label{fig:ex2}
    \end{subfigure}
    
    \begin{subfigure}{\textwidth}
        \centering
        \includegraphics[width=0.8\textwidth]{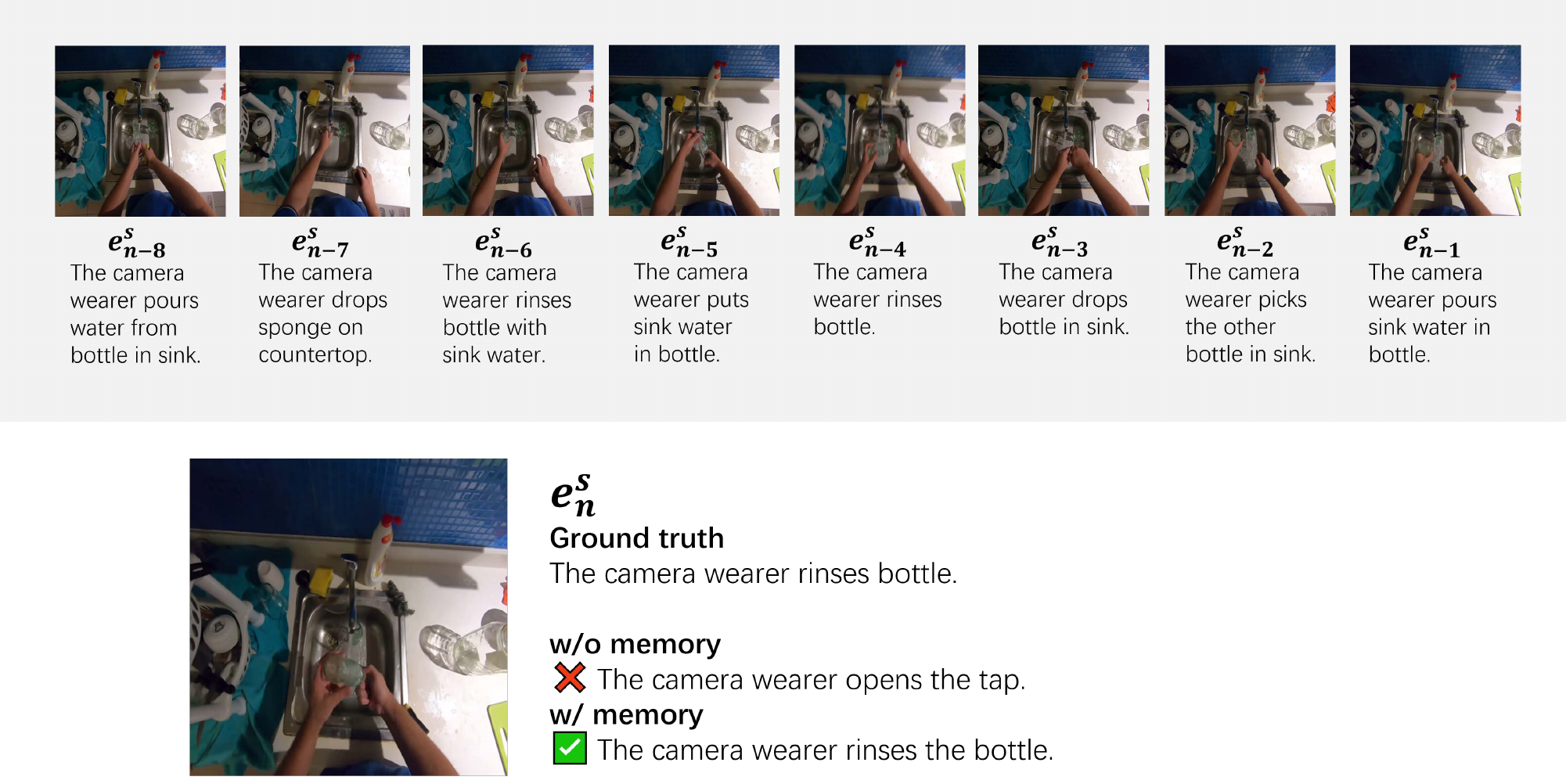}
        \caption{Example 3}
        \label{fig:ex2}
    \end{subfigure}

    \caption{Evaluation of \opt with different ratios and metrics.}
    \label{fig:example}
\end{figure*}

We showcase three examples as in Fig.~\ref{fig:example}. They demonstrate that leveraging short-term memory as contexts can significantly help to understand the current event. 

\end{document}